\documentclass{article}
\usepackage{arxiv}
\usepackage{amsmath}
\usepackage{mathtools}
\usepackage{graphicx}
\usepackage{colortbl}
\definecolor{Gray}{gray}{0.8}

\usepackage[section]{placeins}
\usepackage{hyperref}
\usepackage{caption}
\usepackage{subcaption}
\usepackage{changepage}
\usepackage{makecell}
\usepackage{listings}
\usepackage{float}

\usepackage{todonotes}

\usepackage{amssymb}
\usepackage{pifont}
\newcommand{\cmark}{\ding{51}}%
\newcommand{\xmark}{\ding{55}}%

\usepackage[normalem]{ulem}

\renewcommand{\vec}[1]{\mathbf{#1}}

\makeatletter
\AtBeginDocument{%
  \expandafter\renewcommand\expandafter\subsection\expandafter
    {\expandafter\@fb@secFB\subsection}%
  \renewcommand\@fb@secFB{\FloatBarrier
    \gdef\@fb@afterHHook{\@fb@topbarrier \gdef\@fb@afterHHook{}}}%
  \g@addto@macro\@afterheading{\@fb@afterHHook}%
  \gdef\@fb@afterHHook{}%
}
\makeatother

\makeatletter
\AtBeginDocument{%
  \expandafter\renewcommand\expandafter\subsubsection\expandafter
    {\expandafter\@fb@subsecFB\subsubsection}%
  \newcommand\@fb@subsecFB{\FloatBarrier
    \gdef\@fb@afterHHook{\@fb@topbarrier \gdef\@fb@afterHHook{}}}%
  \g@addto@macro\@afterheading{\@fb@afterHHook}%
  \gdef\@fb@afterHHook{}%
}
\makeatother

\title{Some of the variables, some of the parameters, \\ some of the times, with some physics known: \\ Identification with partial information} 

\author{
  Saurabh Malani\\
  The Department of Chemical and Biological Engineering \\
  Princeton University
  \AND
  Tom Bertalan\\
  Department of Chemical and Biomolecular Engineering\\
  Whiting School of Engineering, Johns Hopkins University
  \AND
  Tianqi Cui\\
  Department of Chemical and Biomolecular Engineering\\
  Whiting School of Engineering, Johns Hopkins University\\
  \AND
  Jos\'e L. Avalos\\
  The Department of Chemical and Biological Engineering, Andlinger Center for Energy and the Environment\\ Princeton Bioengineering Initiative, Department of Molecular Biology, High Meadows Environmental Institute\\Princeton University \\
  \AND
  Michael Betenbaugh\\
  Department of Chemical and Biomolecular Engineering\\
  Whiting School of Engineering, Johns Hopkins University\\
  \AND
  Ioannis G. Kevrekidis*\\
  Department of Chemical and Biomolecular Engineering, Department of Applied Mathematics\\
  Whiting School of Engineering, Johns Hopkins University\\
}

\begin{document}

\maketitle



\begin{abstract}
Experimental data is often comprised of variables measured independently, at different sampling rates (non-uniform $\Delta t$ between successive measurements); and at a specific time point only a subset of all variables may be sampled. 
Approaches to identifying dynamical systems from such data typically use interpolation, imputation or subsampling to reorganize or modify the training data {\em prior} to learning. 
Partial physical knowledge may also be available \textit{a priori} (accurately or approximately), and data-driven techniques can complement this knowledge.
Here we exploit neural network architectures based on numerical integration methods and \textit{a priori} physical knowledge to identify the right-hand side of the underlying governing differential equations. 
Iterates of such neural-network models allow for learning from data sampled at arbitrary time points \textit{without} data modification. 
Importantly, we integrate the network with available partial physical knowledge in ``physics informed gray-boxes"; this enables learning unknown kinetic rates or microbial growth functions while simultaneously estimating experimental parameters.\\
\end{abstract}

\clearpage

\section{Introduction}
%
Dynamical systems arise in all areas of science and engineering, from spatiotemporally chaotic or turbulent flows to transient chemical reactor dynamics, complex biological systems, infectious disease models and beyond. These dynamical systems obey fundamental physical or mathematical governing laws, certain components of which may not be known \textit{a priori}, or may only be known approximately: we may have a known functional form but miss certain parameter values (e.g. kinetic constants); or alternatively, we may not even exactly know the right functional form (e.g. the right kinetics).  
System identification is the science of using experimental data to extract information/knowledge about the underlying dynamical system or possibly to deduce a useful surrogate model; such models can be used for prediction, extrapolation of the system behavior beyond the training data range, optimization of its performance, or for controlling it to a desired setpoint.
%
%
This is of course an established and rich branch of systems theory, with many successful applications; yet the recent explosion in machine learning algorithms has caused a resurgence of interest in several issues arising when learning models from time-series data---including the lack of interpretability, the ability to successfully generalize, and dealing with missing data and partial observations.

Many applications of neural ordinary differential equations (Neural ODEs) and Recurrent Neural Networks (RNNs) focus on learning functionals of entire time-series sequences, e.g. classification of ECG signals to heartbeat types \cite{Saadatnejad2020, Yildirim2018}. 
%
%
In this paper our focus is on learning autonomous dynamical systems from partial information, although our approach can be extended to non-autonomous systems. 

Before starting, we state here our definitions for certain terms as used in this work: 
(a) \textbf{Full Observations}: At each sampling time, \textit{all} relevant dependent variables are measured;
(b) \textbf{Partial Observations}: At each sampling time, \textit{a subset of} the variables is measured;
(c) \textbf{Hidden Variables}: some variables are \textit{never} measured at \textit{any} time;
(d) \textbf{Data $\Delta t$} is the time interval between successive sampling times;
(e) \textbf{Learning $\delta t$} is our estimate of an integration time step long enough to be useful, but also short enough to be accurate; 
(f) \textbf{Fixed Frequency Sampling}: Variables are sampled at a consistent sampling frequency (Data $\Delta t$ is fixed); and
(g) \textbf{Variable Frequency Sampling}: Variables are sampled at arbitrary sampling frequencies (Data $\Delta t$ is variable).


System identification is traditionally a data-driven process---using mathematical, statistical or machine learning techniques on input-output data to extract information about the underlying system laws. This branch of systems theory has a long history, from the celebrated Kalman filter \cite{Kalman1960} (a linear method); non-linear variants such as the extended Kalman filter \cite{Ljung1979}, the unscented Kalman filter \cite{Wan2000}, and the particle filter \cite{Gordon1993}; to new techniques such as neural networks \cite{Nelles2001a, Wang2017, Kuschewski1993, Chen1990, Hudson1990, Rico-Martinez1992}.  
%
Empirical data however, while being an essential component of the task, is not the only source of information typically available about the systems of interest. Governing equations, conservation laws, symmetries, physical theories and/or models, even when imperfect or incomplete, also constitute crucial pieces of information. 
In our work here we endeavor to take into account all such sources of information, developing a hybrid gray-box model that combines black-box data driven techniques with white-box physics models.

Related  approaches are becoming increasingly successful (and popular in recent literature), for example through Physics-Informed Neural Networks (PINNs) \cite{Raissi2017} where physical knowledge, often in the form of partial differential equations \cite{Raissi2019}---possibly with unknown parameters, or even with unknown operators \cite{Lu2021}---is enforced while using deep-learning techniques.
However, PINNs are typically used to find solutions consistent with a fully known underlying governing ODE/PDE; our approach centers around using known physically relevant solutions (experimental data) to identify (parts of) the underlying system itself. While the term ``PINN" has come to refer to the forwards (problem solution) task, simultaneous work \cite{Raissi2017} also exhibited some features of the inverse identification task, albeit  without gray-box structure. This was then followed by much more general operator-learning work \cite{Lu2021}, in which general functional operators are learned, including integration; older versions of this operator-learning taks can be found for example in \cite{Gonzalez-Garcia1998}.
%
Performing such a ``black box" task using machine learning, and, in particular, neural networks is a research direction that started in the late 1980s and early 1990s \cite{Farber1993, Rico-Martinez1992, Rico-Martinez1995}, and has been rejuvenated and much more widely used with the advent of neural ODEs or ODE-Nets \cite{Chen2018, Krishnapriyan2022}.

When partial information about the right-hand-side of the underlying differential equations is known, the  ``gray-box" version of this identification process uses the expressivity of neural networks to {\em correct for incompleteness or inaccuracies} of partially/approximately known physical models, or to impose additional structure or known global laws or constraints.
These  corrections can be performed in an additive (e.g. \cite{Rico-Martinez1994, Lee2022, Kemeth2022, Menesklou2021, Thompson1994, Shi2019}) or multiplicative (e.g. \cite{Lovelett2020, Chen2000}) manner, or the neural network can be integrated with physics in a functional manner (e.g. \cite{Kemeth2022, Psichogios1992, Hagge2017, DeVeaux1999, Oliveira2004, VanCan1997}). 

In this paper we will focus on identifying both missing parameters and missing functional terms: known governing  principles such as mass/energy balances or couplings are imposed by the formulation, and the black-box neural network learns physically relevant functions such as microbial growth rates or chemical reaction rates, and simultaneously estimates possibly unknown parameter values.
Note that this is in contrast to methods such as Hamiltonian Neural Networks \cite{Bertalan2019,Greydanus2019}
where a structural feature of the dynamics (symplecticity, or  their Hamiltonian nature) is imposed {\em in addition} to the demand for accurate prediction.

In placing our work in the context of current literature, we consider distinct categories of analogous efforts: first those that are focusing on black-box models trained on data with missing or partial information; second on black-box models trained on data at variable sampling rates;  and last, those that are focusing on partially known physics (gray-box models).\\

\noindent{\bf Black Box Models trained on data with missing or partial information.}
Measurements of experimental data often are not available in a format ideal for data-driven prediction/learning. Sensors may not operate at regular intervals, manually sampled data may also not be available at regular intervals, and some data points may be corrupted by external noise. This results in a dataset that is sampled at \textit{variable frequencies}. Furthermore, different sensors may operate at different sampling rates; and at each sample point only some analyses may be run. This results in a dataset with \textit{partial observations}---where at each sample point, only a subset of all variables are measured.

To make such training data conducive to statistical analyses and machine learning, pre-processing is often performed; this may be as simplistic as omitting the ``anomalous" datapoints \cite{Lyngdoh2022}, but more often some form of imputation is involved. This could can be linear-interpolation or a ``last observation carried forward" approach \cite{Gelman2006, Oluwaseye2022}, may involve more robust smoothing techniques such as polynomial or LOESS smoothing (locally estimated scatterplot smoothing) \cite{Honaker2010}, machine learning techniques such as expectation maximization \cite{Nelwamondo2007a}, k-nearest neighbours \cite{Bertsimas2021, Silva-Ramirez2015, Oluwaseye2022, Lyngdoh2022} or artificial neural networks \cite{Silva-Ramirez2015, Nelwamondo2007a} for smoothing and imputation.

Methodologies that impute missing data \textit{first} and train a model on the cleaned dataset \textit{second} often reinforce the imputed values {\em as if they were true}:
%
``if you fill the holes in the cheese with peanut butter, you should not pretend to have more cheese!'' \cite{Honaker2010}.
This can be mitigated by conducting imputation and model training \textit{simultaneously}: using the model to impute missing data, further training the model using the data (but not including the loss at the imputed (not ground-truth) values); using the improved model to impute again, and iterating.\\

\begin{figure}[h!]
    \centering
    \includegraphics[width=\textwidth]{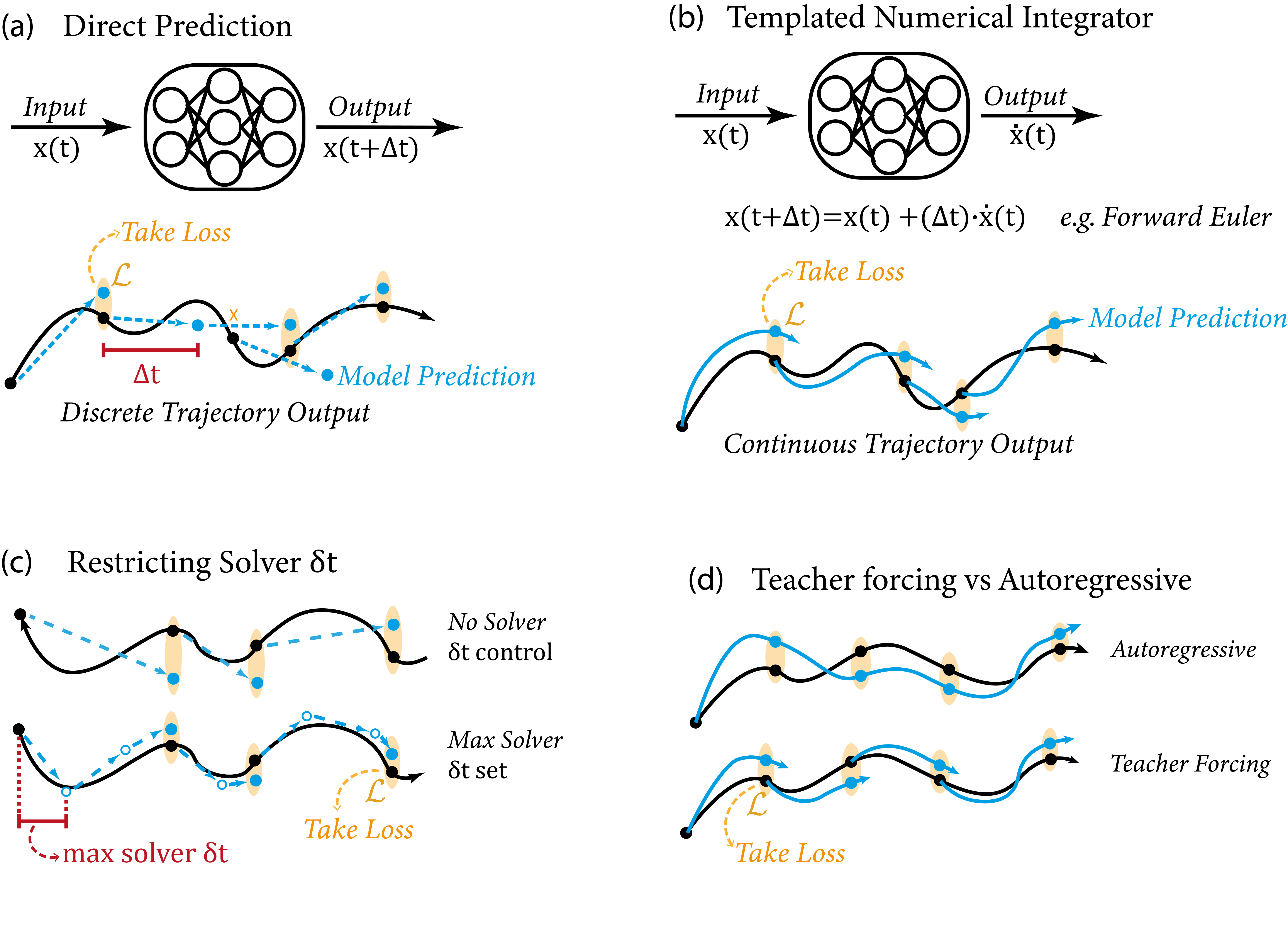}
    \caption[Comparison of different RNN types.]{\textbf{Comparison of different RNN types.} (a) Direct prediction methods typically force the network into a fixed $\Delta t$, which prompts the resampling, interpolation, or imputation of data that don't happen to be sampled at this $\Delta t$. (b) Incorporating the neural network in a numerical integrator template with a $\Delta t$ input naturally allows for prediction at varying $\Delta t$ values. (c) Restricting the max solver $\delta t$ hyperparameter forces network to take multiple steps if data $\Delta t$ is large. (d) We explore both teacher-forcing, where real data is used as input when available, and autoregressive, where predictions are fed back into network for the next prediction. During training on partial observations, elements of both teacher forcing and autoregression are used, while during inference-time, autoregression is used, starting from a known initial condition.
    }
    \label{fig:IntroFig}
\end{figure}

\noindent{\bf Black Box Models trained on data sampled at variable time points.} To train on data sampled at variable sampled time points, we need a method that can naturally allow for arbitrary time stepping. The simplest supervised learning approach involves learning a direct mapping between the current state $x(t)$ and the state $x(t+\Delta t)$ at a time $\Delta t$ in the future, but this method either requires training on data that is sampled at regular $\Delta t$ intervals, or requires the $\Delta t$ itself also be an input of the network (Fig \ref{fig:IntroFig}(a)). 
An alternative (and in our opinion more flexible) approach consists of recognizing that the underlying systems can be represented as differential equations in continuous time. Hence, if we instead have the neural network learn {\em the form of the right hand side of a differential equation system}, we can leverage well-established numerical integrator methods to generate predictions at any desired time point (Fig \ref{fig:IntroFig}(b)). This method was proposed in the literature as far back as the 1990's \cite{Rico-Martinez1992,Rico-Martinez1993} and has been extended and popularized in recent work through neural ODEs \cite{Chen2018}. This recent work in Neural ODEs has allowed for compatibility with built-in, robust numerical integrators in Python, with gradients calculated using the adjoint method rather than unrolling, and packages such as \textit{torchdiffeq} \cite{Chen2018a} and \textit{diffrax} \cite{Kidger2022} have become publicly available. 

Neural ODEs also provide a natural way of simultaneously imputing and learning, where the imputation is done with the time-evolution model and dynamically changes in every iteration as the model is trained. In this work, we ``open up" the Neural ODE approach by explictly templating the neural network on a 4th order Runge-Kutta (RK4) integrator in PyTorch \cite{Rico-Martinez1992} (Fig \ref{fig:RK4Fig}). In our work the training gradients are computed by unrolling, rather than through the adjoint method.\\

\noindent{\bf Gray Box Models --- Learning from data with some physics known}.
Various previous works in the literature have touched on different combinations of missing data characteristics. Rubanova et. al. use a neural ODE
to learn on irregularly-sampled time series data \cite{Rubanova2019}, for data with completely hidden variables---but they do not consider partial observations. They operate on a latent representation that makes gray-boxing difficult, since latent variables need not have straightforward physical interpretations.
%
%
%

Kidger et al. \cite{Kidger2020} extend neural ODEs to neural controlled differential equations, where the hidden state evolves continuously with time. They demonstrate the neural CDE methodology on a variety of classification tasks, but don't deal with system identification from time-series or integration with hybrid gray-box methods. Instead, this application represents a direct continuous-time analog of the typical functional-learning RNN application.

Buisson-Fenet et. al. \cite{Buisson-Fenet2022} give a simple demonstration of integrating neural ODEs with prior knowledge to produce a particular type of hybrid gray-box methods for system identification; here the physical knowledge is given in the form of known kinetic equations
while the neural network learns a system  parameter (the unknown frequency). 
%
In addition to working with a spectrum of gray models by incorporating prior information, a key contribution of \cite{Buisson-Fenet2022} is the use of a synchronized dynamical system based on the (Kazantzis-Kravaris-Luenberger) KKL observer theory to impute missing initial conditions. 
This addresses training on data with hidden variables, but not with partial observations, as in the current work.
Approaches using kernel methods---rather then neural networks---have also been proposed \cite{Bouvrie2017, Hamzi2021}) for learning dynamical systems or for predicting missing initial conditions of LSTM latent dynamics \cite{Kemeth2021}.

In our work, the neural network learns physically relevant functions such as microbial growth rates or a chemical reaction rates alongside unknown values of fixed experimental parameters.

The remainder of the paper is organized as follows: Having provided the definitions of the terms we use above, we start by summarizing the key concepts underpinning our approach and the algorithm.  We demonstrate our approach first on a well-established illustrative model---the autonomous stirred tank reactor \cite{Uppal1974}; where we incrementally build on the complexity of the training data used to learn the model. 
We then proceed to a more complex system---a 6-dimensional biological kinetics model involving three microbial species growing in co-culture and exhibiting autonomous oscillations \cite{BALTZIS1984}. 
In both systems, we demonstrate how gray-boxing may be used to both estimate unknown parameters and  simultaneously learn unknown dynamics.
We conclude with a summary of our observations, and a discussion of the scope of the approach and of its strengths and weaknesses.

\section{Methodology}

\begin{figure}[h!]
    \centering
    \includegraphics[width=0.5\textwidth]{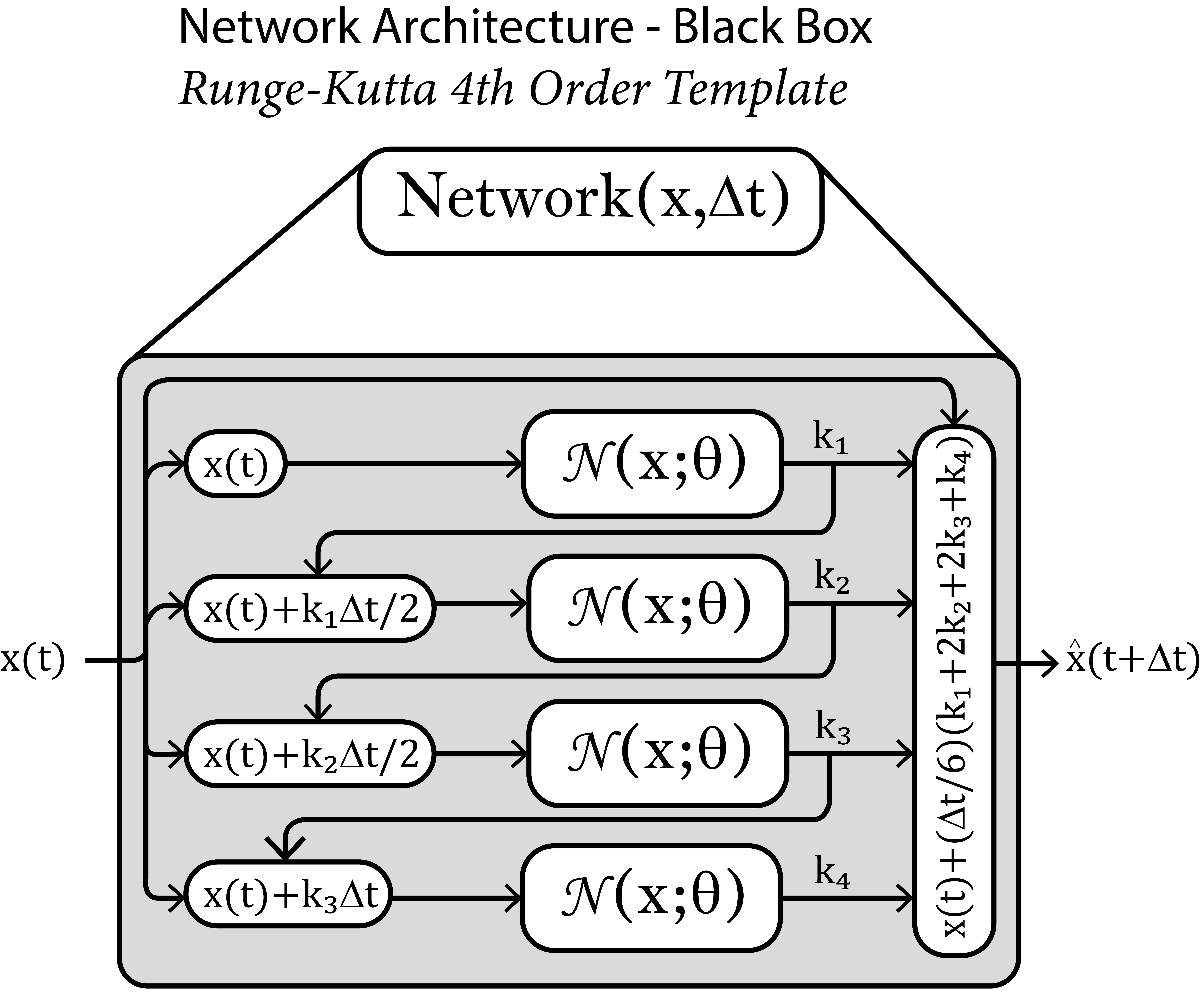}
    \caption[Runge-Kutta network]{A black box neural network model with a fourth order Runge-Kutta numerical integrator}
    \label{fig:RK4Fig}
\end{figure}

\textbf{ODE-nets for learning with arbitrary time sampling.} In order for the neural network to learn the Right-Hand-Side (RHS) of a system of ODEs (rather than learning to directly predict the output in discrete time), it is convenient if the network architecture has been templated on an established numerical integrator framework (Fig \ref{fig:IntroFig} (a), Fig \ref{fig:IntroFig} (b)) as a ``fusion'' of neural networks with traditional numerical analysis which we call ODE-nets (but may also be referred to in literature as Neural ODEs \cite{Chen2018}). In this work we explictly template the neural network on a 4th order Runge-Kutta (RK4) (Eq \ref{eq:RK4}, Fig \ref{fig:RK4Fig}).

\begin{subequations}
\begin{equation}
    \hat{\dot{\mathbf{x}}} = \mathcal{N}(\mathbf{x}(t),\mathbf{p};\theta)
\end{equation}
\begin{equation}
    \begin{aligned}
        \vec{k}_1 &= \mathcal{N}(\mathbf{x}(t),\mathbf{p};\theta)\\
        \vec{k}_2 &= \mathcal{N}(\mathbf{x}(t)+\vec{k}_1\left(\frac{\Delta t}{2}\right),\mathbf{p};\theta)\\
        \vec{k}_3 &= \mathcal{N}(\mathbf{x}(t)+\vec{k}_2\left(\frac{\Delta t}{2}\right),\mathbf{p};\theta)\\
        \vec{k}_4 &= \mathcal{N}(\mathbf{x}(t)+\vec{k}_3\left(\Delta t\right),\mathbf{p};\theta)\\
        \mathbf{\hat{x}}(t+\Delta t) &= \mathbf{x}(t) + \left(\frac{\Delta t}{6}\right)(\vec{k}_1 + 2\vec{k}_2 + 2\vec{k}_3 + \vec{k}_4)
    \end{aligned}
\end{equation}
\label{eq:RK4}
\end{subequations}

We know from numerical analysis that, generally speaking, the forward error in integrating a known equation improves as step size decreases, motivating the development of adaptive-stepsize integration schemes. Likewise, it can be shown that the backwards error, in discovering an equation from assumed-true trajectory data, also improves as step size decreases \cite{Zhu2022}. 
As such, we control the time steps the RK4 integrator takes in order to maintain the precision of the solver, such that multiple iterations through the network may be needed to obtain a prediction at the next observation time step (Fig \ref{fig:IntroFig} (c)).

\begin{figure}[h!]
    \centering
    \includegraphics[width=\textwidth]{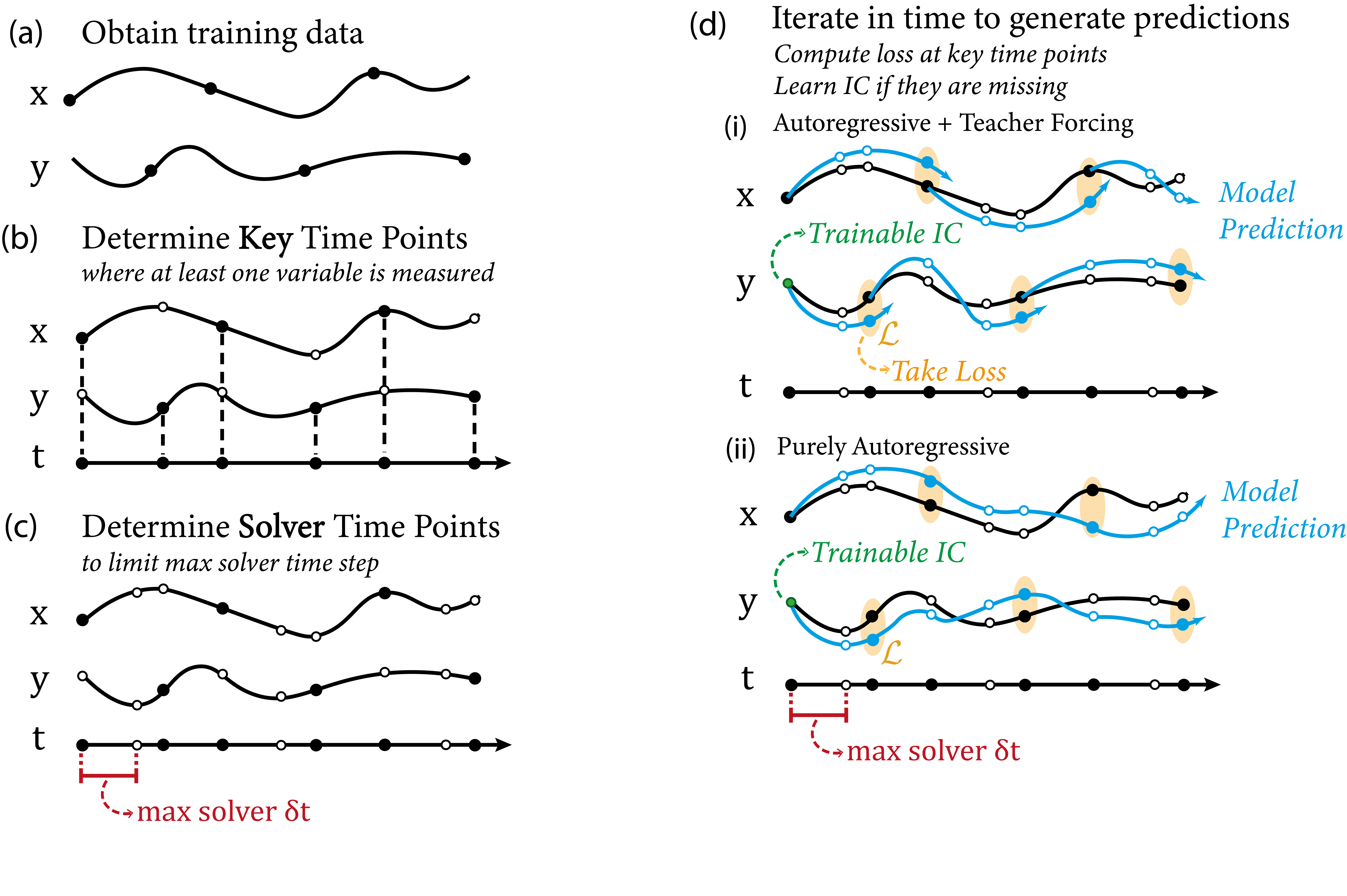}
    \caption[Training Algorithm and Batching.]{\textbf{Training Algorithm and Batching.} (a) Systematically record each trajectory of training data. (b) Determine the \textbf{key} time points for each trajectory; key time points are where at least one variable is measured and hence a loss will be computed at these time points. (c) Determine \textbf{solver} time points. No loss is computed at these time points, these are included to ensure stability and accuracy of the numerical integrator. (d) Generate predictions by integrating through each trajectory in time; (i) teacher-forcing is used when real data is available for a feature, and if not predictions are generated autoregressively. In this paper this approach is used during \textit{model training}; (ii) model predictions are generated purely autoregressively, starting from a given initial condition. This paper uses this approach during \textit{inference} time. For training, compute loss at the key-time points for each trajectory, back-propagate to compute gradients and perform gradient descent to train the network.}
    \label{fig:MethodFig}
\end{figure}

We also note two differing approaches to training: {\em teacher-forcing} and {\em autoregressive} (Fig \ref{fig:IntroFig} (d)): teacher forcing was first introduced in the context of recurrent networks for temporal supervised learning \cite{Williams1989, Pineda1988} and involves replacing the output of the network with the true data (teacher signal), when available, for subsequent iterations. This approach allows for faster convergence of the model during training, particularly during earlier iterations \cite{BennyToomarian1992}. In contrast, autoregressive training involves feeding back the outputs of the network back into itself as the input for the next iteration to yield a full temporal trajectory \cite{Chen2018}. Previous literature has suggested that a balance between the two is optimal for training and model predictions \cite{BennyToomarian1992}, and we take this into account in our training methodology (Fig \ref{fig:MethodFig}).\\

\noindent\textbf{Gray-Boxing for data-driven learning with some physics known.}  We aim to develop and implement a neural-network based training methodology amenable to coupling with known, ``white-box'', physical knowledge, and capable of learning from training data sets with variable sampling times and partial observations. 
Both of these requirements can be addressed by formulating the training problem as the learning of the RHS of a system of ODEs.
Often, established physical knowledge is expressed in terms of differential equations in space and time (chemical or microbial kinetics, convection, diffusion, etc.)
Expressing the dynamics in the form of differential equations imposes the right inductive bias during the neural network model training to capture the continuous nature of these dynamical systems \cite{Krishnapriyan2022}; and it naturally allows the neural network training to incorporate existing partial physical knowledge. The methods can be extended to PDEs and even SDE/SPDEs \cite{Dietrich2021, Psarellis2022}, but that is beyond the scope of this paper

\begin{figure}[h!]
    \centering
    \includegraphics[width=0.8\textwidth]{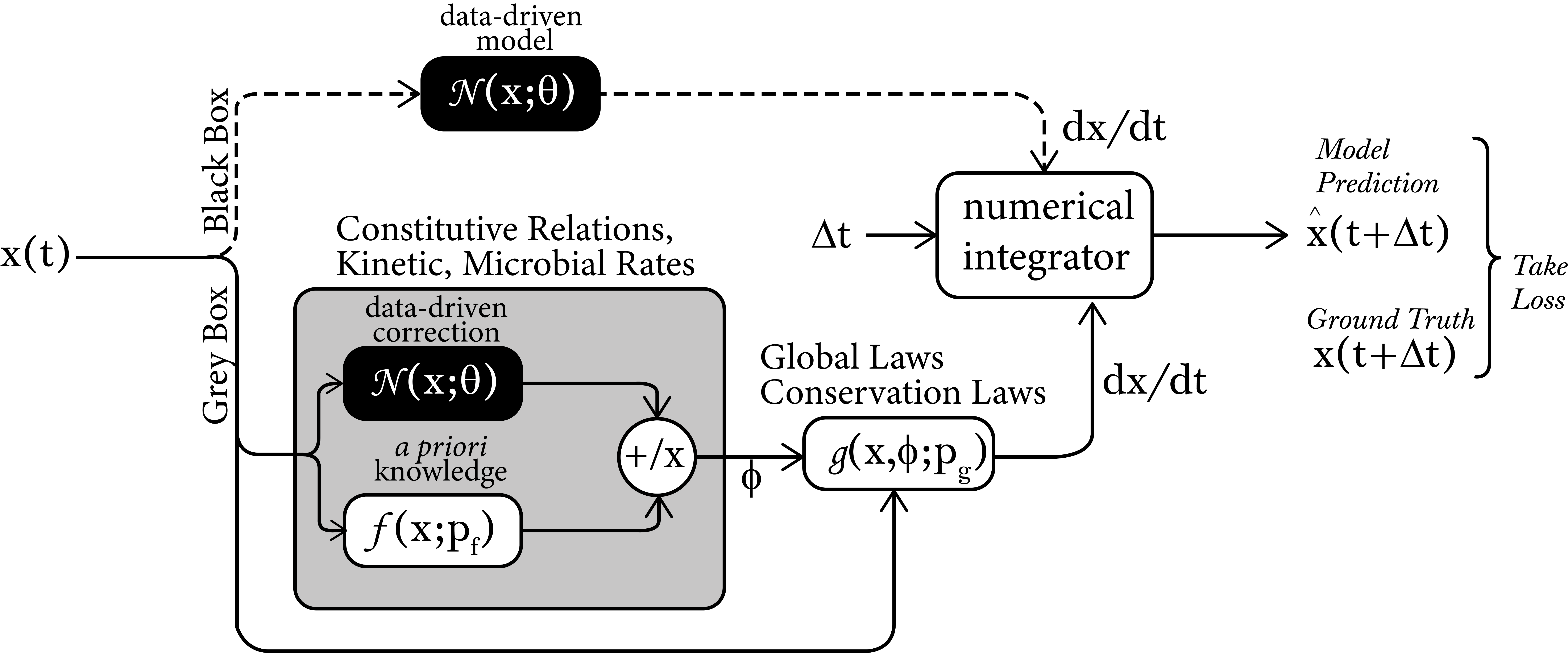}
    \caption[Network architecture for black and gray box models.]{\textbf{Network architecture for black and gray box models.} In the black box model, the entire onus for learning the RHS is on the neural network. In the gray-box model, physical information is incorporated through constitutive terms or known microbial and reaction rates, and global and conservation laws.}
    \label{fig:GrayBoxFig}
\end{figure}

In a simple black-box model, the neural network will have the full onus of learning the law governing the system dynamics:
\begin{subequations}
    \begin{equation}
        \hat{\dot{\vec{x}}} = \mathcal{N}_{BB}(\vec{x};\theta)
    \end{equation}\\
    Previously available physical knowledge, which may be incomplete and/or imperfect, may be included in the model framework, so that the neural network is tasked with learning only the unknown parameters/dynamics, or possibly a correction to imperfectly known dynamics. This inclusion can be corrective (either additive (Eq \ref{eq:GB_add}) or multiplicative (Eq \ref{eq:GB_mult})); or it could be a function composition (Eq \ref{eq:GB_func}).\\
    \begin{equation}
        \hat{\dot{\vec{x}}} = f_{WB}(\vec{x};\vec{p}) + \mathcal{N}_{BB_{Add}}(\vec{x};\theta)
        \label{eq:GB_add}
    \end{equation}
    \begin{equation}
        \hat{\dot{\vec{x}}} = f_{WB}(\vec{x};\vec{p}) \cdot \mathcal{N}_{BB_{Mult}}(\vec{x};\theta)
        \label{eq:GB_mult}
    \end{equation}
    \begin{equation}
        \hat{\dot{\vec{x}}} = f_{WB}(\vec{x},\mathcal{N}_{BB_{Func}}(\vec{x});\theta)
        \label{eq:GB_func}
    \end{equation}
\end{subequations}
\begin{subequations}
    We can also combine corrective gray-boxes and function composition gray-boxes (Eq \ref{eq:GB_full}, Fig \ref{fig:GrayBoxFig})
    \begin{equation}
    \begin{aligned}
        \phi = f_\phi(\vec{x};\vec{p_f},\theta) = \begin{cases}
            f_{WB}(\vec{x};\vec{p_f}) + \mathcal{N}_{BB_{Add}}(\vec{x};\theta) & \text{Additive}\\
            f_{WB}(\vec{x};\vec{p_f}) \cdot \mathcal{N}_{BB_{Mult}}(\vec{x};\theta) & \text{Multiplicative}\\
        \end{cases}
    \end{aligned}
    \end{equation}
    \begin{equation}
        \hat{\dot{\vec{x}}} = g_{WB}(\vec{x},\phi;\vec{p_g})
    \end{equation}
    \label{eq:GB_full}
\end{subequations}
where $f_\phi$ are constitutive relations, or functions modeling kinetic or microbial rates, for which we may have \textit{a priori} postulates  $f_{WB}$. The neural network learns a correction for any imperfections or incompleteness in $f_{WB}$, or learns the entire $f_\phi$ if there are no \textit{a priori} postulates. $\phi$ and $x$ are inputs to another white box model $g_{WB}$ which strictly imposes inviolable physical laws, such as conservation laws or other global laws we may want to impose, such as a coupling between reaction rate and the heat of reaction. Note that here, the white box parameters $\vec{p_f}$ and $\vec{p_g}$ can either be fixed \textit{a priori}, or, if sufficiently identifiable given the data, they can be included as additional training parameters to be fit simultaneously with the learning of the neural network parameters $\theta$.

\begin{subequations}
    In our approach, we focus on function composition, where the neural network handles the task of learning the unknown microbial or kinetic rates, and the architecture consists of a white-box structure imposing known
    physical couplings (e.g. between reaction rate and reactive heat generation, or microbial growth and substrate consumption/product formation (Eq \ref{eq:GB_ours}).
    \begin{equation}
        \phi = \mathcal{N}_{BB}(x;\theta)\\
    \end{equation}
    \begin{equation}
        \hat{\dot{x}} = g_{WB}(x,\phi;p_g)
    \end{equation}
    \label{eq:GB_ours}
\end{subequations}
\paragraph{Training Algorithm and Batching.} The training algorithm is summarized in Figure \ref{fig:MethodFig}. Within each training trajectory, \textbf{key} time points (where the loss is calculated), and \textbf{solver} time points (included for stability and accuracy of the templated numerical integrator) are marked, and predictions are generated (with the current vectorfield estimate) by iterating in time through each trajectory.
Each training trajectory may have different key and solver time points, and may also be of differing length. To include multiple trajectories in the same training minibatch, a sequence of evaluation time points is included alongside the data input, and padding is added to make all data and time sequences the same length. This allows for simultaneous training on minibatches of data trajectories.\\

\noindent\textbf{Loss function and error metrics.} For the loss function, a mean squared error (MSE) loss is taken between the model predicted trajectory and the ground truth data at the time points at which data is available:

\begin{subequations}
    \begin{equation}
    L = \frac{1}{N_\mathrm{data}} \sum_{i\in\mathrm{trajectories}}\sum_{j\in\mathrm{times}_i}\sum_{k\in\mathrm{channels}(i,j)}(\hat{y}_i^{(k)}(t_j) - y_i^{(k)}(t_j))^2.\\
    \end{equation}
    \begin{equation}
    N_\mathrm{data}=\sum_{i\in\mathrm{trajectories}}\sum_{j\in\mathrm{times}_i}||\mathrm{channels}(i,j)||
    \end{equation}
\end{subequations}
Here, $\mathrm{channels} (i,j)$ is the set of channel indices observed in trajectory $i$ at times $j$.
Refer to Section \ref{sec:Metrics} for details on metrics used to evaluate model performance.

\section{Results}

\begin{figure}[h!]
    \centering
    \begin{subfigure}[b]{0.75\textwidth}
        \includegraphics[width=\textwidth]{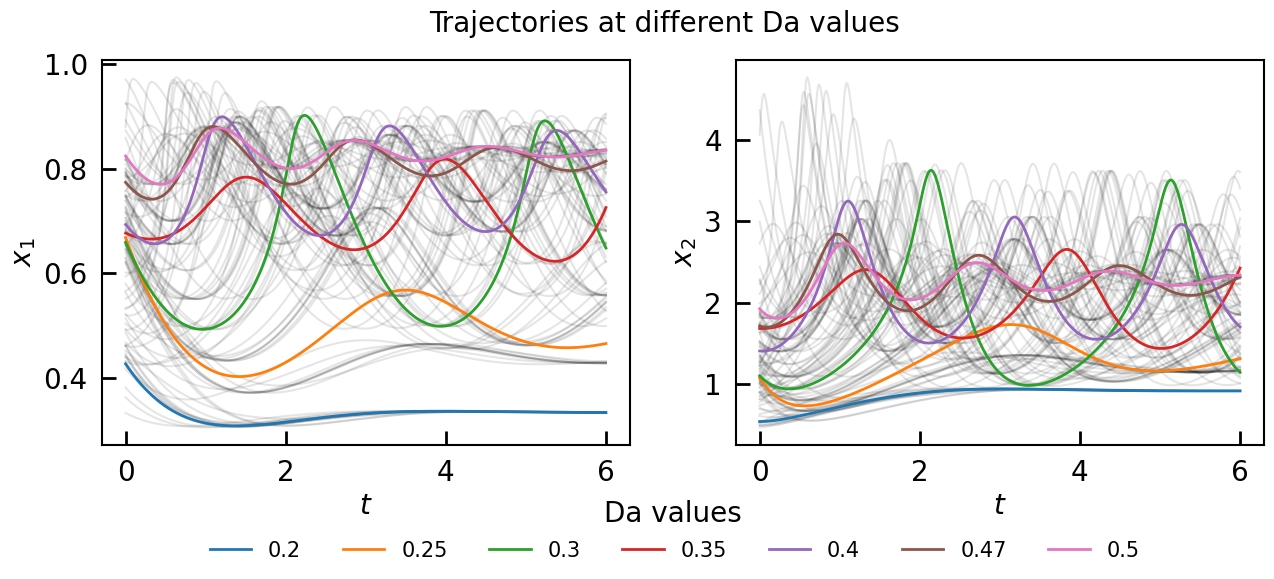}
        \caption{Training data trajectories at various $Da$ values}
    \end{subfigure}
    \begin{subfigure}[b]{0.75\textwidth}
        \includegraphics[width=\textwidth]{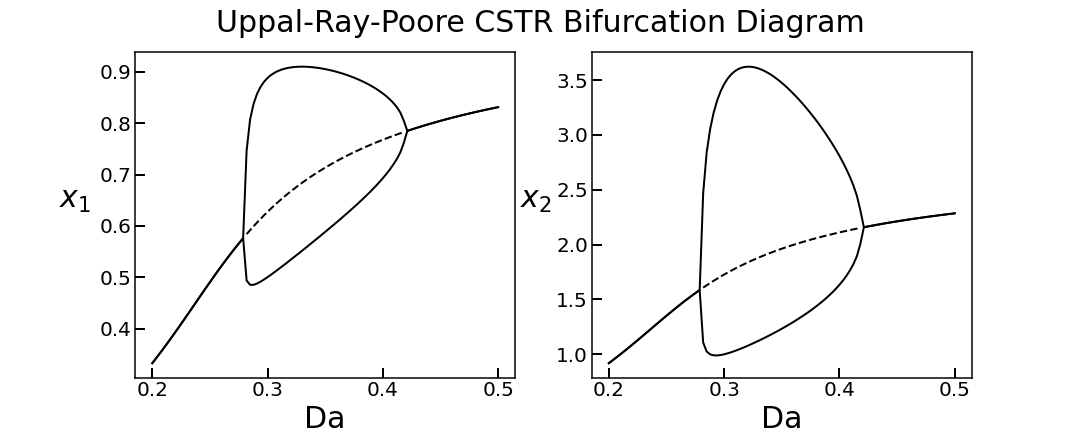}
        \caption{Bifurcation Points at $Da=0.280275$ and $Da=0.419548$. \\Solid lines: Stable steady state \textit{or} max/min values of stable oscillations;\\ Dashed line: Unstable steady state}
    \end{subfigure}
    \caption{Visualization of Training data for URP CSTR}
    \label{fig:URP_TrainingData}
\end{figure}

Our first illustrative example uses data from simulations of the non-isothermal CSTR model in the classical chemical engineering literature \cite{Uppal1974} (URP CSTR Model), a two dimensional system of ordinary differential equations (Eq \ref{eq:URPModel}),

\begin{subequations}
    \begin{equation}
        \cfrac{dx_1}{dt} = -x_1 + \mathrm{Da}\cdot (1-x_1)\cdot \exp{(x_2)}=f_1(x_1,x_2)
    \end{equation}
    \begin{equation}
        \cfrac{dx_2}{dt} = -x_2 + \mathrm{B}\cdot \mathrm{Da}\cdot (1-x_1)\cdot \exp{(x_2)} - \mathrm{\beta} \cdot x_2 = f_2(x_1,x_2)
    \end{equation}
where $x_1$ is non-dimensional concentration (conversion), $x_2$ is non-dimensional temperature, and $t$ is non-dimensional time.
    \label{eq:URPModel}
\end{subequations}
This system is parameterized by the fixed parameters $B=11$ and $\beta=3.0$, and a variable parameter $Da$ (the Damkohler number) sampled in the range $\mathrm{Da}\in(0.2,0.5)$.
For this choice of $B$ and $\beta$, the model exhibits two Hopf bifurcation points wrt. Da: at $\mathrm{Da}=0.280275$ and $\mathrm{Da}=0.419548$ respectively (Fig \ref{fig:URP_TrainingData}).
Stable periodic oscillatory behavior is observed in the range $0.280275\leq \mathrm{Da}\leq 0.419548$, and a single stable steady state outside of this range. Sample trajectories of training data are visualized in Figure \ref{fig:URP_TrainingData}.

The neural network architecture here is a multi-layer perceptron with the following structure: (a) a three-neuron input layer with SiLU activation; (b) two hidden layers of 64 neurons each with SiLU activation; and (c) an output layer with two neurons (black-box formulation) or a single neuron (gray-box formulation) with linear activation.

In the ``Black-Box'' formulation, no \textit{a priori} knowledge of the system is assumed, and so the neural network must  learn the \textit{full} dynamics (Eq \ref{eq:URP_BB}) and its dependence on $Da$.
\begin{equation}
    \hat{\dot{\mathbf{x}}} = \mathcal{N}(\mathbf{x},\mathrm{Da};\theta)\;\;\;\text{where } \mathbf{x}=\{x_1,x_2\}
    \label{eq:URP_BB}
\end{equation}
In the gray-box formulation, {\em some} parts of the system dynamics are known \textit{a priori}, while others are unknown (Eq \ref{eq:URP_GB}). Here, the kinetics of the reaction are assumed to be unknown, but the form of the overall mass and energy conservation laws, as well as the coupling of the reaction rate to reactive heat generation (the heat of reaction, which can be measured independently) are assumed to be known.
\begin{subequations}
\begin{equation}
    \begin{aligned}
    \hat{\phi} &= \mathcal{N}(\mathbf{x},Da;\theta)\\
    \hat{\dot{x_1}} &= -x_1 + \hat{\phi}\\
    \hat{\dot{x_2}} &= -x_2 + \textcolor{blue}{\mathrm{B}}\cdot \hat{\phi} - \textcolor{blue}{\mathrm{\beta}} \cdot x_2
    \end{aligned}
\end{equation}\\
Here $\hat{\phi}(\mathbf{x},Da)$ is the neural network approximation of the reaction kinetics (Eq \ref{Eq:URP_GB_Phi}).
\begin{equation}
    \phi = f_\phi(\vec{x},\mathrm{Da}) = \mathrm{Da}\cdot (1-x_1)\cdot \exp{(x_2)}
    \label{Eq:URP_GB_Phi}
\end{equation}
Here $\textcolor{blue}{\mathrm{B}}$ and $\textcolor{blue}{\mathrm{\beta}}$ are experimental parameters that can be independently measured, are assumed  known \textit{a priori} and are ``hardwired" in the network; if they happen to {\em not} be known, they can of course be considered as \textit{additional} training parameters, to be fitted along with $Da$ and the neural network weights during ANN training.

For simplicity, unless otherwise specified,  $\mathcal{N}$ represents the overall network trained to predict $\hat{\dot{\mathbf{x}}}$, with all gray-box aspects subsumed within this representation.
\begin{equation}
    \hat{\dot{\mathbf{x}}} = \mathcal{N}(\mathbf{x},\mathrm{Da};\theta)
\end{equation}
\label{eq:URP_GB}
\end{subequations}
\begin{table}[H]
\centering
\resizebox{1.0\textwidth}{!}{%
\begin{tabular}{ |c|ccccccc| }
\hline
\rowcolor{Gray}
\textbf{Case} & \textbf{A} & \textbf{B} & \textbf{C} & \textbf{D} & \textbf{E} & \textbf{F} & \textbf{URP\_BB}\\
\hline
Data $<\Delta t>$ & 0.1 & 0.5 & 0.5 & 0.5 & \makecell{0.5 ($x_1$)\\ 0.55($x_2$)} & 0.5 & 0.5 \\
\hline
\makecell{Solver \\ max $\delta t$} & 0.1 & 0.5 & 0.1 & 0.1 & 0.1 & 0.1 & 0.1\\
\hline
\makecell{Arbitrary \\Data $\Delta t$} & - & - & - & \cmark & - & \cmark & \cmark\\
\hline
\makecell{ Partial \\ Observations} & - & - & - & - & \cmark & \cmark & \cmark\\
\hline
\makecell{ Initial Condition \\ Given} & \cmark & \cmark & \cmark & \cmark & \cmark & \cmark & -\\
\hline
\makecell{Solution \\ Error ($\mathcal{L}_2$)} & \makecell{$(9.68\pm 1.71)$\\$\times 10^{-3}$} 
& \makecell{$(1.49\pm 0.12)$\\$\times 10^{-2}$} & \makecell{$(8.69\pm 1.18)$\\$\times 10^{-3}$} 
& \makecell{$(4.23\pm 0.43)$\\$\times 10^{-3}$} & \makecell{$(6.09\pm 0.56)$\\$\times 10^{-3}$}
& \makecell{$(8.91\pm 1.65)$\\$\times 10^{-3}$} & \makecell{$(7.92\pm 1.36)$\\$\times 10^{-3}$}\\
\hline
\makecell{RHS \\ Error ($\mathcal{L}_2$)} & \makecell{$(2.72\pm 0.36)$\\$\times 10^{-3}$}
& \makecell{$(9.71\pm 0.76)$\\$\times 10^{-3}$} & \makecell{$(5.95\pm 1.21)$\\$\times 10^{-3}$}
& \makecell{$(3.50\pm 0.34)$\\$\times 10^{-3}$} & \makecell{$(3.66\pm 0.46)$\\$\times 10^{-3}$}
& \makecell{$(7.12\pm 0.81)$\\$\times 10^{-3}$} & \makecell{$(7.69\pm 0.11)$\\$\times 10^{-3}$}\\
\hline
\end{tabular}}
\caption{Metrics Summary Table for URP CSTR training. This Table showcases model performance on data of increasing complexity of representation. Errors are means and standard deviations of 10 training runs. Refer to Section \ref{sec:Metrics} for details on error metrics.}
\label{tab:UPR_Metrics_Cases}
\end{table}

\begin{table}[h!]
\centering
\resizebox{1.0\textwidth}{!}{%
\begin{tabular}{ |>{\columncolor[gray]{0.8}}c|cccccc| } 
 \hline
 \textbf{Case} & \makecell{Learnable \\ Kinetic \\Functions} & \makecell{Learnable \\ Experimental \\Parameters} & \makecell{ Solution \\ Error ($\mathcal{L}_2$)} & \makecell{ RHS \\ Error ($\mathcal{L}_2$)}  & \makecell{ Kinetic \\Function \\ Error ($\mathcal{L}_2$)} & \makecell{ Experimental \\Parameter \\ Error ($\mathcal{L}_2$)}\\ 
 \hline
 \textbf{URP\textunderscore BB}& - & - &
 $(7.92\pm 1.36)\times 10^{-3}$
 & 
$(7.69\pm 0.11)\times 10^{-3}$
 &-&-\\ 
 \textbf{URP\textunderscore GB1}& \cmark & - &
$(9.91\pm 1.26)\times 10^{-3}$
 & 
$(4.43\pm 0.51)\times 10^{-3}$
 &
$(5.31\pm0.62) \times 10^{-3}$
 &-\\ 
 \textbf{URP\textunderscore GB2}& \cmark & \cmark &
$(1.22\pm 0.17)\times 10^{-2}$
 & 
$(6.49\pm 2.08)\times 10^{-3}$
 &
$(7.77\pm 2.25) \times 10^{-3}$
 &
 $(4.45\pm 2.28)\times 10^{-2}$
 \\ 
 \hline
\end{tabular}}

\caption{Metrics Summary Table for URP CSTR data. This table showcases model performance of Black-box and Gray-box models. All models have arbitrary data $\Delta t$, partial observations and no initial condition given. Errors are means and standard deviations of 10 training runs. Refer to Section \ref{sec:Metrics} for details on error metrics.}
\label{tab:UPR_Metrics_BBGB}
\end{table}

\begin{figure}[h!]
    \centering
    \begin{subfigure}[t]{0.4\textwidth}
      \centering
      \includegraphics[width=\textwidth]{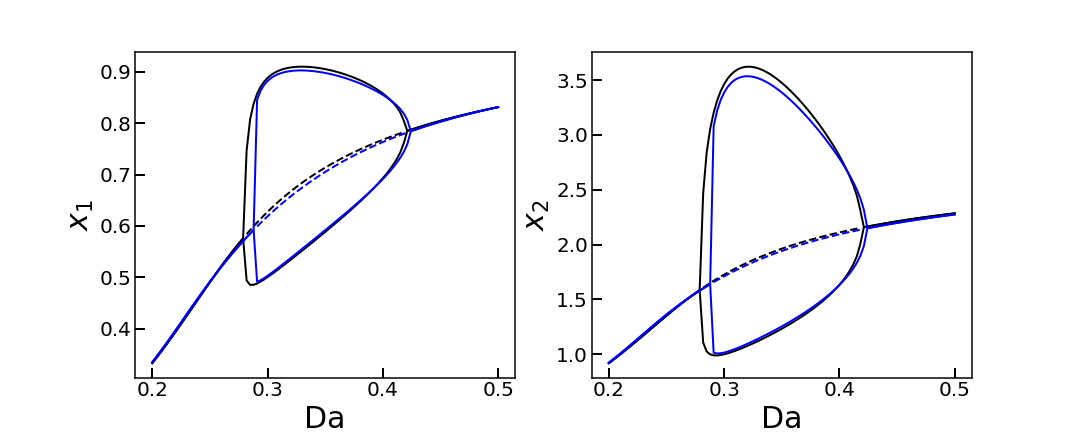}
      \caption{Case A: Base Case: Ideal data (Finely sampled, full observations, fixed time sampling)}
      \label{fig:URP_Bifurc_CaseA}
    \end{subfigure}
    \hfill
    \begin{subfigure}[t]{0.45\textwidth}
      \centering
      \includegraphics[width=\textwidth]{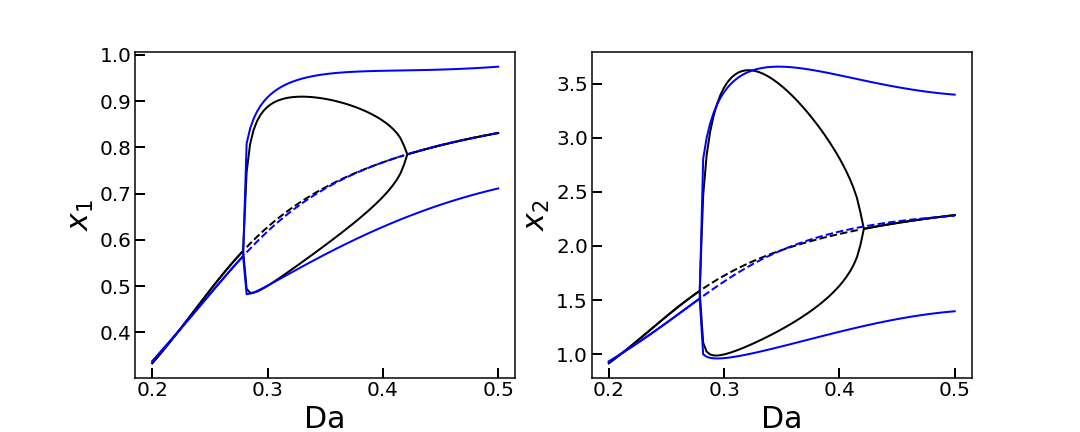}
      \caption{Case B: Training with no solver time step control}
      \label{fig:URP_Bifurc_CaseB}
    \end{subfigure}
    \begin{subfigure}[t]{0.45\textwidth}
      \centering
      \includegraphics[width=\textwidth]{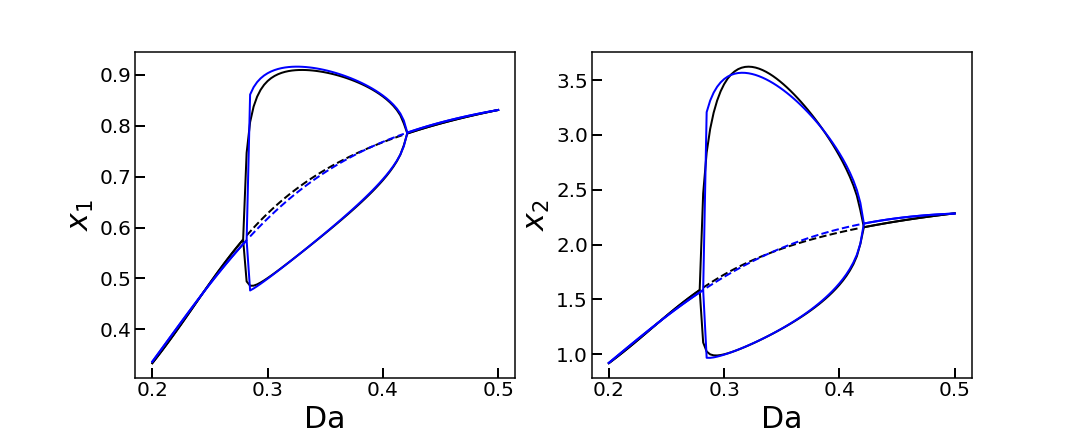}
      \caption{Case C: Training with solver time step control}
      \label{fig:URP_Bifurc_CaseC}
    \end{subfigure}
    \hfill
    \begin{subfigure}[t]{0.45\textwidth}
      \centering
      \includegraphics[width=\textwidth]{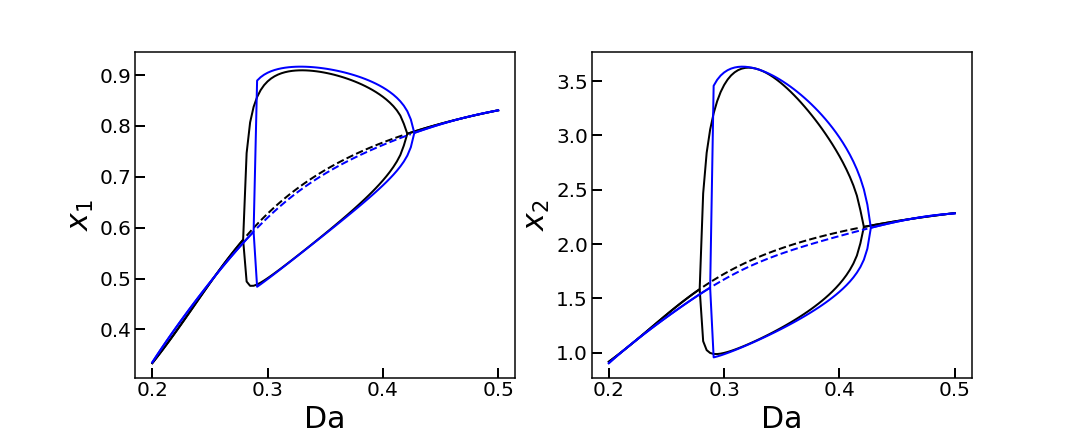}
      \caption{Case URP\_BB: Final Model (Solver time step control, partial observations, variable time step sampling, trainable initial conditions}
      \label{fig:URP_Bifurc_CaseBB}
    \end{subfigure}
    \caption{Bifurcation Diagram Predictions}
    \label{fig:URP_Bifurc}
\end{figure}

In our demonstration we gradually ``build up" the complexity in the training data used. All model results are summarized in Tables \ref{tab:UPR_Metrics_Cases} and \ref{tab:UPR_Metrics_BBGB}. Case A (Fig \ref{fig:URP_Bifurc_CaseA}) is the base case, with the training data trajectories densely sampled in time, with full observations and regular time sampling. For this simple base case, we observe good reproduction of the bifurcation diagram by the black box trained network.
Cases B (Fig \ref{fig:URP_Bifurc_CaseB}) and C (Fig \ref{fig:URP_Bifurc_CaseC}) demonstrate the importance of controlling the solver max $\delta t$. In Case B, the solver max $\delta t=0.5$ is too large for stable/accurate integration with the embedded RK4 integrator, and the model is unable to reproduce the Hopf Bifurcation point near $Da=0.42$. Reducing the solver max $\delta$ to $0.1$ in Case C resolves this issue.
In Cases D and E, we introduce arbitrary time sampling and partial observations respectively; Case F has both of these data pathologies simultaneously. In all three we observe good training results with low solution and RHS error. Finally, in Case URP\_BB (Fig \ref{fig:URP_Bifurc_CaseBB}), we let the initial condition vector be trainable, allowing for training on data where the initial condition is not given/where not all variables are measured at the first sampling time.

\begin{figure}[h!]
    \centering
    \begin{subfigure}[b]{0.65\textwidth}
      \centering
      \includegraphics[width=\textwidth]{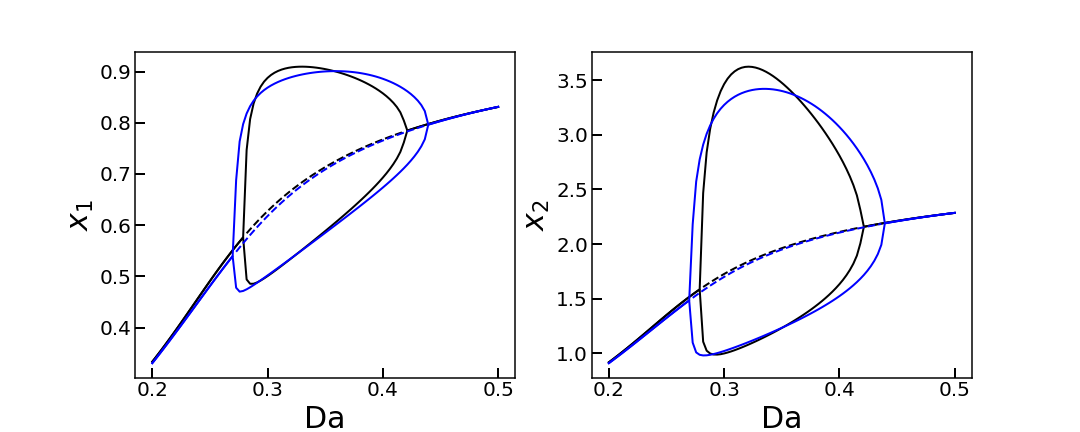}
      \caption{Bifurcation Diagram}
    \end{subfigure}
    \begin{subfigure}[b]{0.3\textwidth}
      \centering
      \includegraphics[width=\textwidth]{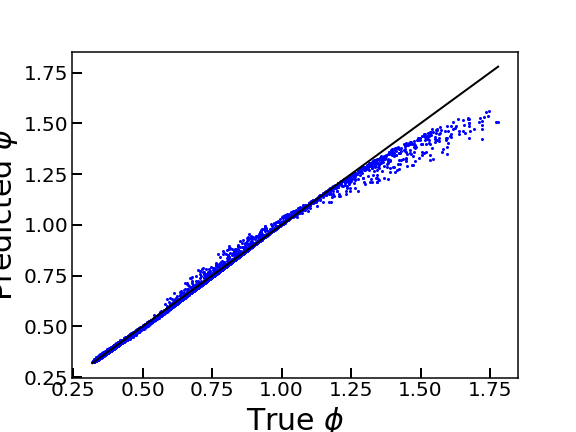}
      \caption{Prediction of \\Kinetic Term}
    \end{subfigure}
    \begin{subfigure}[b]{0.65\textwidth}
      \centering
      \includegraphics[width=\textwidth]{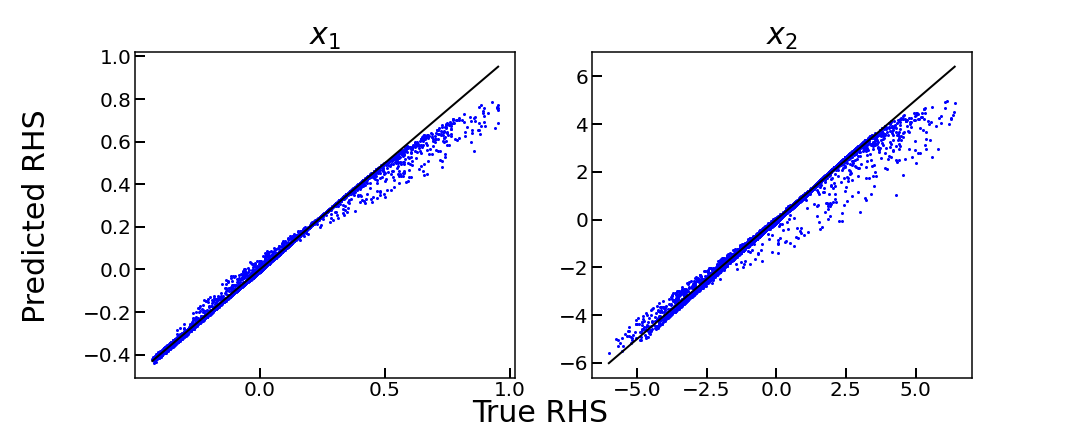}
      \caption{Prediction of RHS}
    \end{subfigure}
    \begin{subfigure}[b]{0.3\textwidth}
      \centering
      \includegraphics[width=\textwidth]{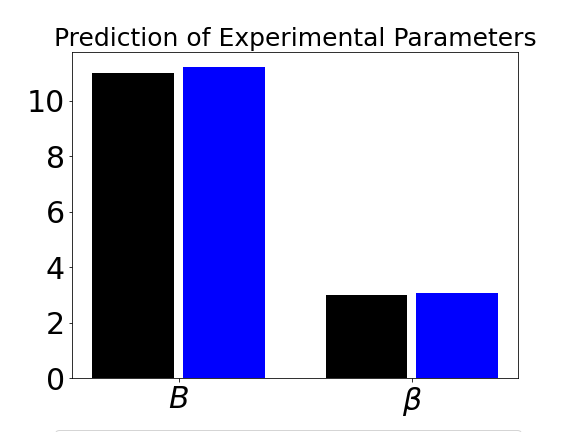}
      \caption{Prediction of \\Experimental Parameters}
    \end{subfigure}
    \begin{subfigure}[b]{0.6\textwidth}
      \centering
      \includegraphics[width=\textwidth]{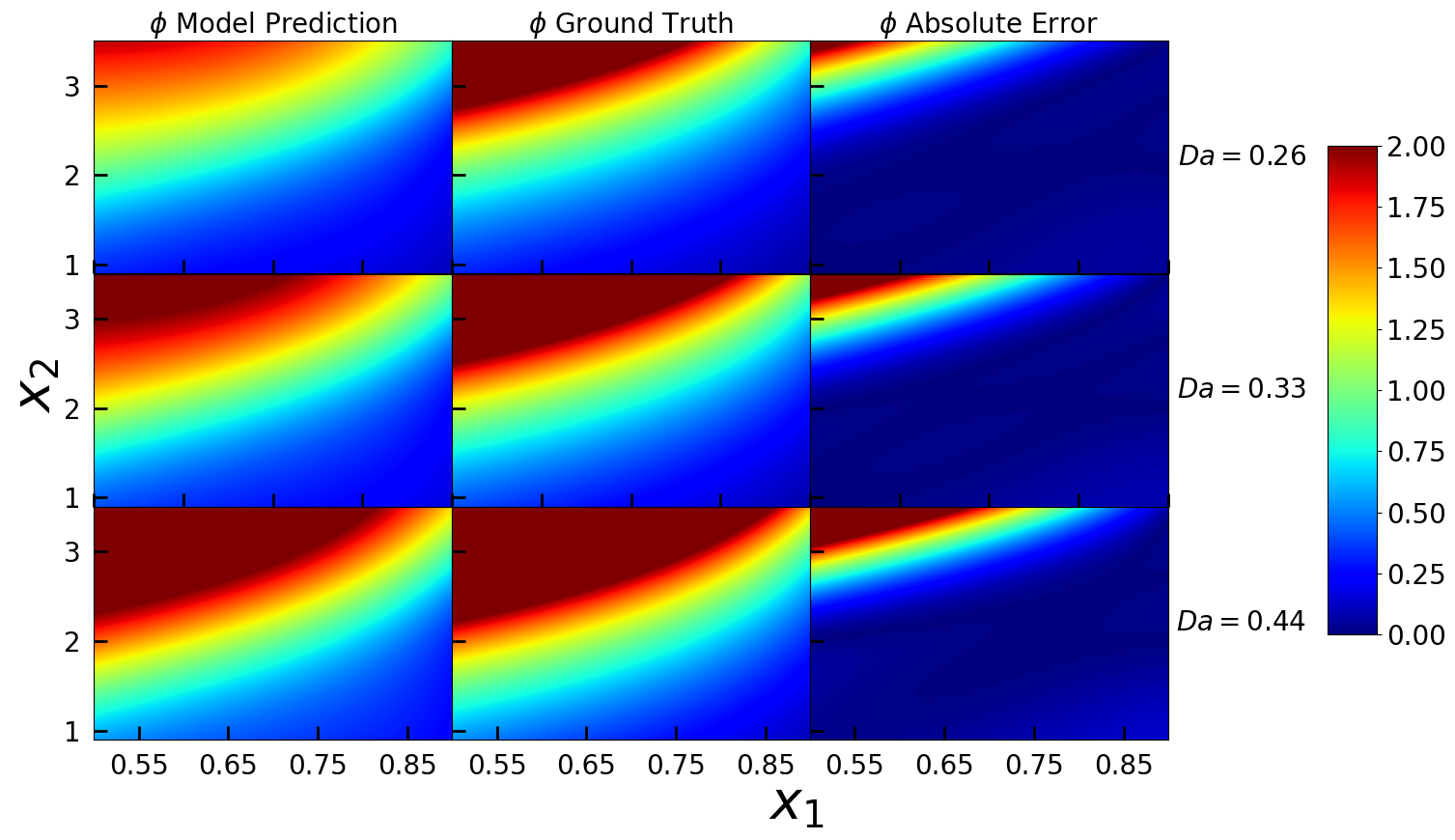}
      \caption{Heatmap for learned $\phi$ and predicted $\phi$ for various $Da$ values}
    \end{subfigure}
    \caption{Trained model performance for Case URP\_GB2 (Gray-Box with trainable experimental parameters)}
    \label{fig:URP_GB2}
\end{figure}

In Cases URP\_GB1 and URP\_GB2 (Fig \ref{fig:URP_GB2}, Table \ref{tab:UPR_Metrics_BBGB}) we introduce physics informed ``gray-boxes" into our model. In both, rather than learning the dynamic evolution of variables $x_1$ and $x_2$ directly, the model instead learns the kinetic function $f\phi$; the evolution of $x_1$ and $x_2$ is reconstructed by mass and energy balances, and coupling of reaction kinetics and energetics. In Case URP\_GB1, the coupling constants $B$ and $\beta$ are known, while in Case URP\_GB2 they are unknown and introduced as additional trainable parameters for the model. In both cases, the data has the pathology of having partial observations, arbitrary time sampling, and no initial condition given. In Case URP\_GB2 (Fig \ref{fig:URP_GB2}) we observe good reproduction of the bifurcation diagram, accurate prediction of the kinetic function $f_\phi$ that translates into accurate RHS predictions for both $x_1$ and $x_2$, and the coupling constants $B$ and $\beta$ are well predicted.

\begin{figure}[h!]
    \centering
    \begin{subfigure}[b]{0.45\textwidth}
      \centering
      \includegraphics[width=\textwidth]{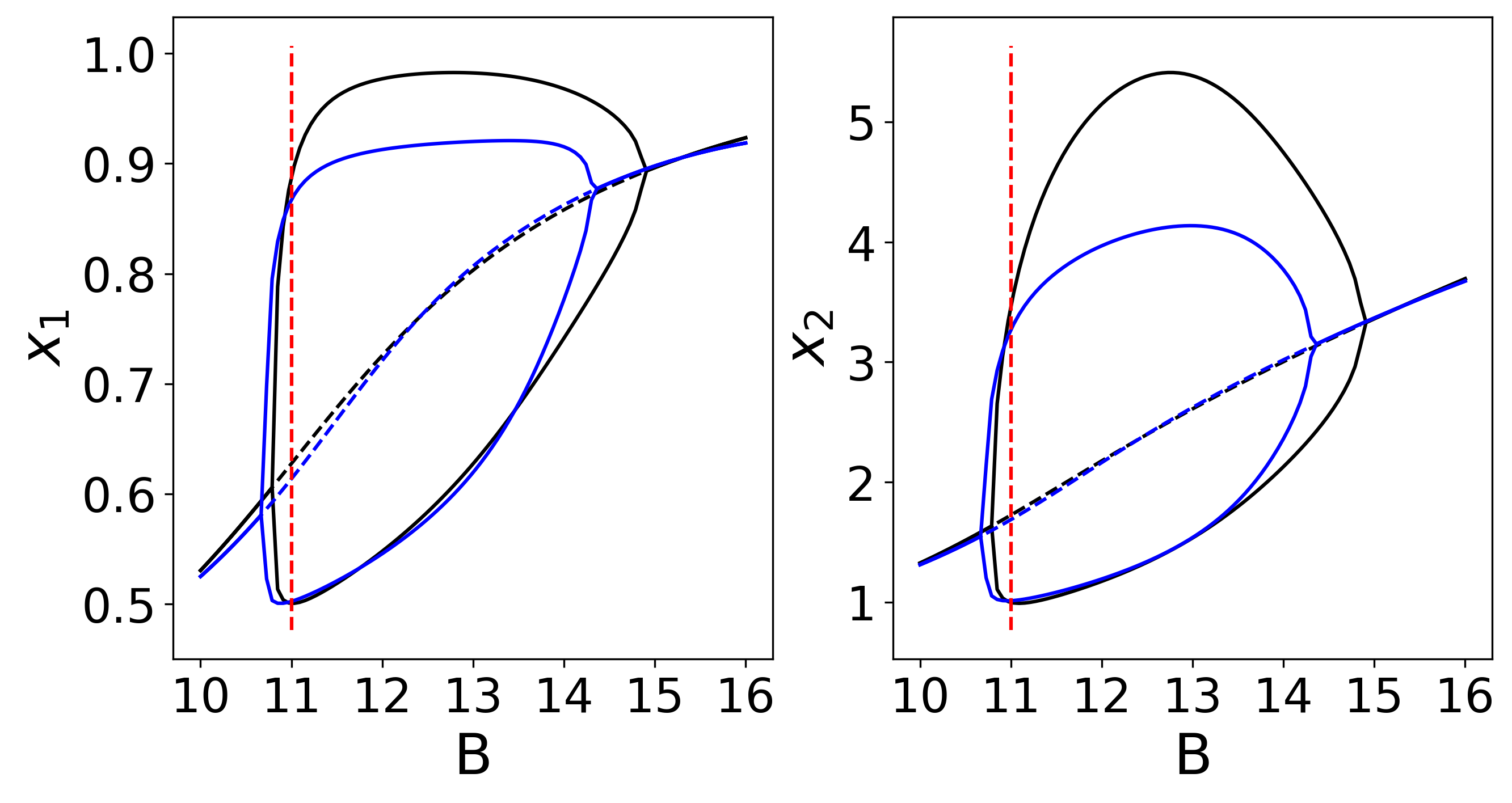}
      \caption{Bifurcation Diagram for B}
    \end{subfigure}
    \centering
    \begin{subfigure}[b]{0.45\textwidth}
      \centering
      \includegraphics[width=\textwidth]{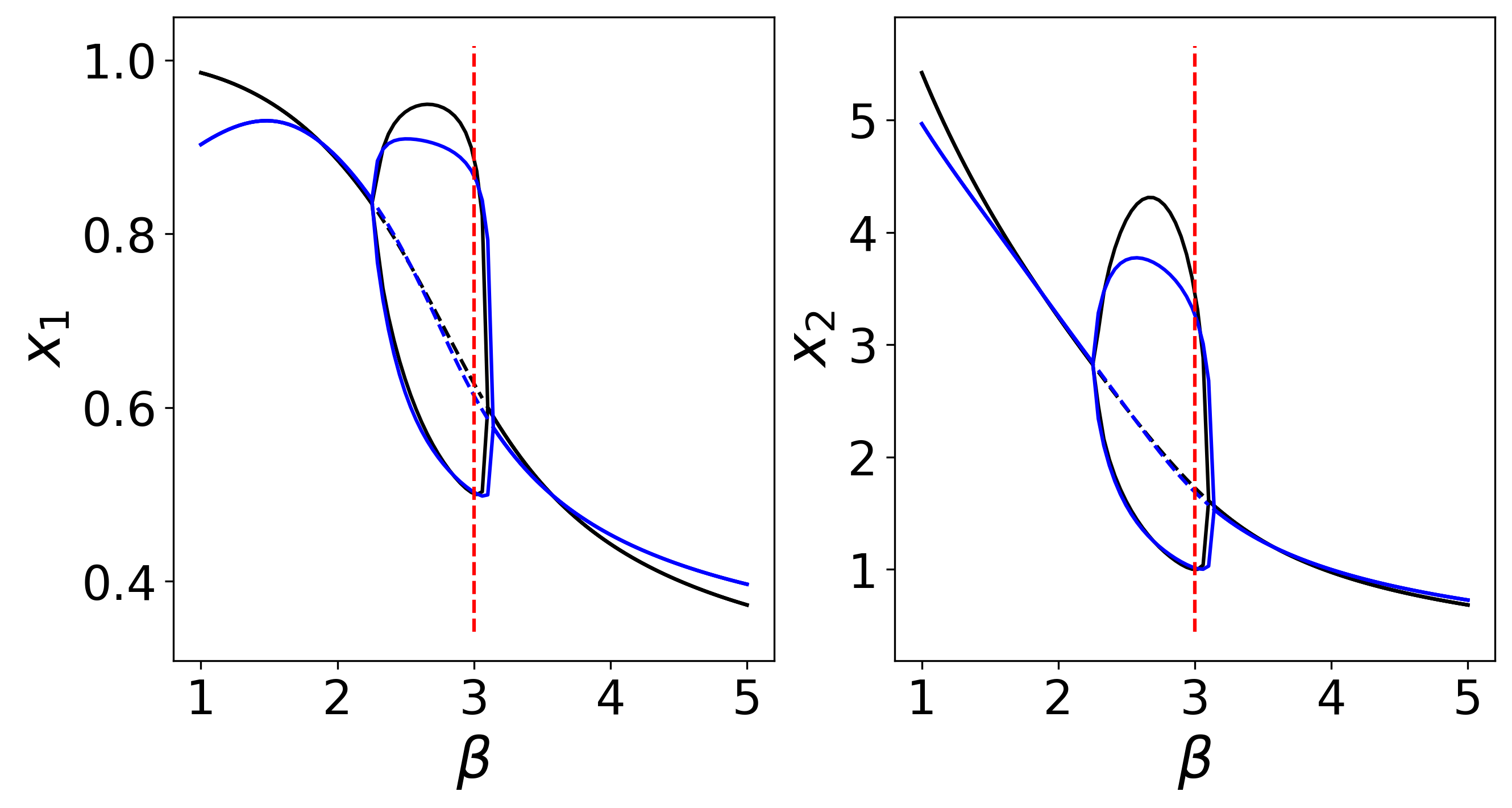}
      \caption{Bifurcation Diagram for $\beta$}
    \end{subfigure}
    \caption{Case URP\_GB1 (Gray-Box with fixed/known experimental parameters): Extrapolating to new parameter values. The dotted red line is at parameter value of training data; all other parameter spaces are extrapolations beyond training data.}
    \label{fig:URP_GB1_Extrapolation}
\end{figure}

A major benefit of training in ``gray-box'' as opposed to ``black-box'' is that it retains physical intuition, and in doing so we expect that it will allow for extrapolation to parameter space beyond the training data. This is evidenced in the Case URP\_GB1 (Fig \ref{fig:URP_GB1_Extrapolation}); this model was trained on a fixed, known value of $B$ and $\beta$, and the neural network was used to learn the kinetic reaction term. This allows the model to be extrapolated to values of $B$ and $\beta$ well beyond those trained on, allowing for predictions of bifurcation diagrams for B and $\beta$ that are remarkably close to ground truth despite the model having been trained at \textit{a single} set of parameter values.

\begin{figure}[h!]
    \centering
    \begin{subfigure}[b]{0.6\textwidth}
        \centering
      \includegraphics[width=\textwidth]{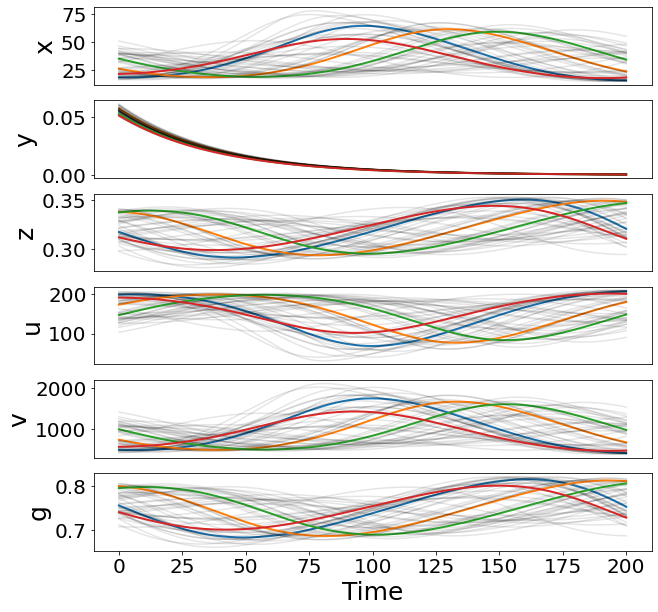}
    \end{subfigure}
    \begin{subfigure}[b]{0.32\textwidth}
        \centering
      \includegraphics[width=\textwidth]{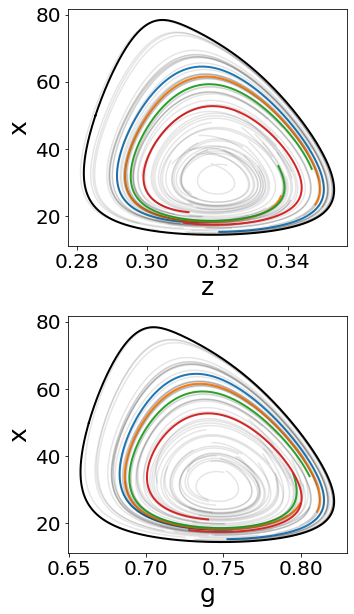}
    \end{subfigure}
    \caption{Transients and Limit Cycles of B\&F training data (200 trajectories with 4 emphasized). Stable Limit Cycle given in Black.}
    \label{fig:BF_trainingdata}
\end{figure}

For our second illustrative example, we use data from a classical biochemical model involving the coexistence of three microbial species growing in co-culture, coupled through various chemical substrates and cofactors \cite{BALTZIS1984} (B\&F Model). This is a six-dimensional system of coupled ODEs outlined in Eq \ref{eq:BFModel}, with parameter values in Table \ref{tab:BFpar}. The training data is visualized in Figure \ref{fig:BF_trainingdata}.
\begin{subequations}    
\begin{equation}
\begin{aligned}
    \cfrac{dx}{dt} &= - \alpha x + \mu_1 x - \mu_{c_1} x && \text{Host Population}\\
    \cfrac{dy}{dt} &= - \alpha y + \mu_2 y - \mu_{c_2} y && \text{Commensal Population 1}\\
    \cfrac{dz}{dt} &= - \alpha z + \mu_3 z - \mu_{c_3} z && \text{Commensal Population 2}\\
    \cfrac{du}{dt} &= \alpha(u_f-u) - \mu_1 x && \text{Host Substrate}\\
    \cfrac{dv}{dt} &= -\alpha v + \omega \mu_1 x - \mu_2 y - \sigma \mu_3 z && \text{Growth Factor}\\
    \cfrac{dg}{dt} &= -\alpha g + \rho \mu_2 y + \eta \mu_3 z  && \text{Inhibition Factor},
\end{aligned}
\end{equation}
where
\begin{equation}
\begin{aligned}
    \mu_1 = f_{\mu_1}(u,g) &= \cfrac{u}{(1+u)(1+g)} && \text{Host Population Growth Rate}\\
    \mu_2 = f_{\mu_2}(v) &= \cfrac{\phi_1 v}{1+v} && \text{Commensal Population 1 Growth Rate}\\
    \mu_3 = f_{\mu_3}(v) &= \cfrac{\phi_2 v}{\sigma + v} && \text{Commensal Population 2 Growth Rate}.
\end{aligned}
\label{eq:BF_true_phi}
\end{equation}
The host $x$ grows on substrate $u$ and secretes growth factor $v$. Commensal populations $y$ and $z$ consume growth factor $v$ and release inhibition factor $g$ that inhibits growth of the host $x$. $f_{\mu_i}$ are the growth rate functions of each species, and $\mu_{ci}$ is a constant parameter dictating maintenance rate of each biomass species.
\label{eq:BFModel}
\end{subequations}
\begin{table}[h!]
\begin{center}
\begin{tabular}{ c | c }
 Parameter & Value \\ 
 \hline
 $\alpha$ & $1/7.3$ \\  
 $u_f$ & $250.0$\\
 $\omega$ & $9.7$\\
 $\sigma$ & $10.0$\\
 $\rho$ & $0.13138686$\\
 $\eta$ & $1.29166$\\
 $\phi_1$ & $0.2941176$\\
 $\phi_2$ & $0.367647$\\
 $\mu_{c_1}$ & $0.367647$\\
 $\mu_{c_2}$ & $0.117647$\\
 $\mu_{c_3}$ & $0.1617647$\\
\end{tabular}
\end{center}
\caption{Parameter Values for Baltzis \& Frederickson (B\&F) Model}
\label{tab:BFpar}
\end{table}

\begin{figure}[h!]
    \centering
    \begin{minipage}[b]{0.45\textwidth}
    \begin{subfigure}[b]{\textwidth}
      \centering
      \includegraphics[width=\textwidth]{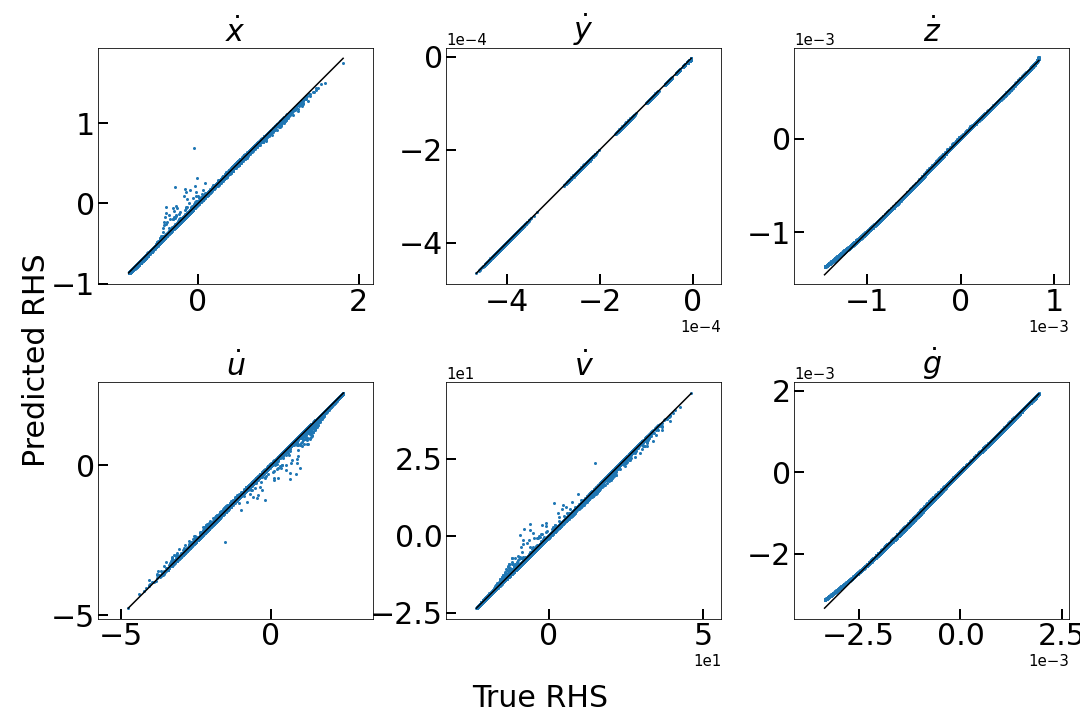}
      \caption{RHS Prediction}
    \end{subfigure}
    \begin{subfigure}[b]{\textwidth}
      \centering
      \includegraphics[width=\textwidth]{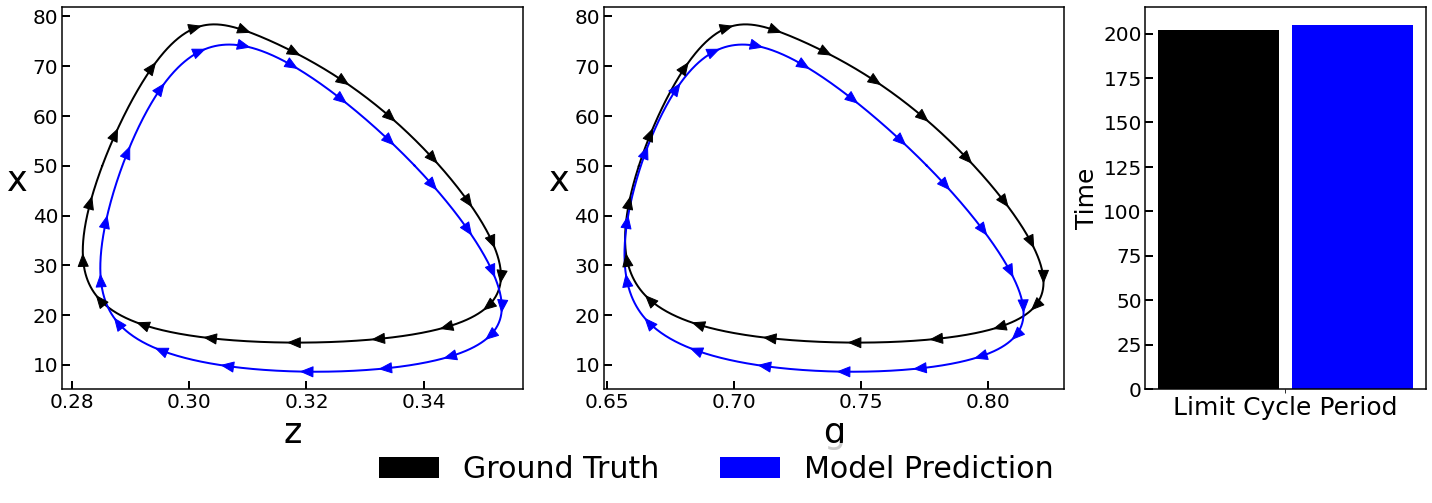}
      \caption{Limit Cycle Prediction}
    \end{subfigure}
    \end{minipage}
    \hfill
    \begin{subfigure}[b]{0.5\textwidth}
      \centering
      \includegraphics[width=\textwidth]{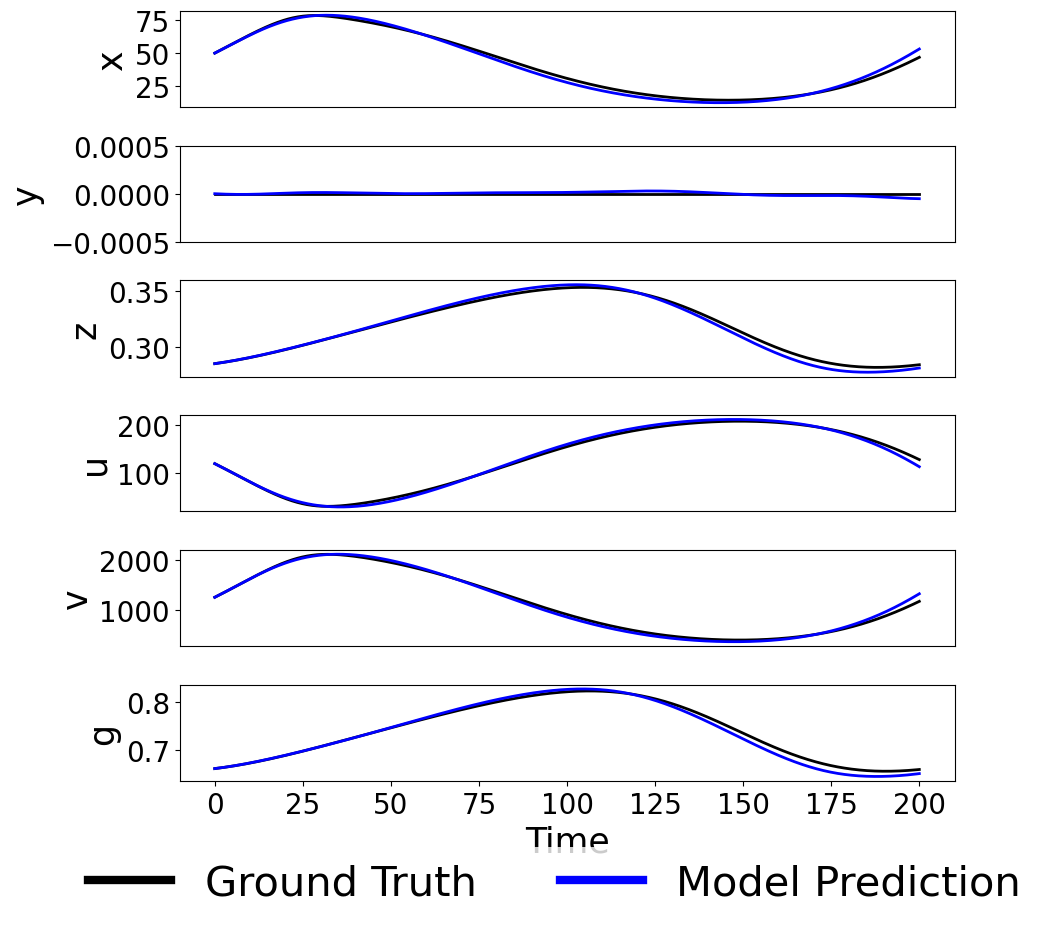}
      \caption{Short Transient Prediction}
    \end{subfigure}
    \caption{Trained model performance for Case BF\_BB}
    \label{fig:BF_BB}
\end{figure}

\begin{figure}[h!]
    \centering
    \begin{minipage}[b]{0.45\textwidth}
    \begin{subfigure}[b]{\textwidth}
      \centering
      \includegraphics[width=\textwidth]{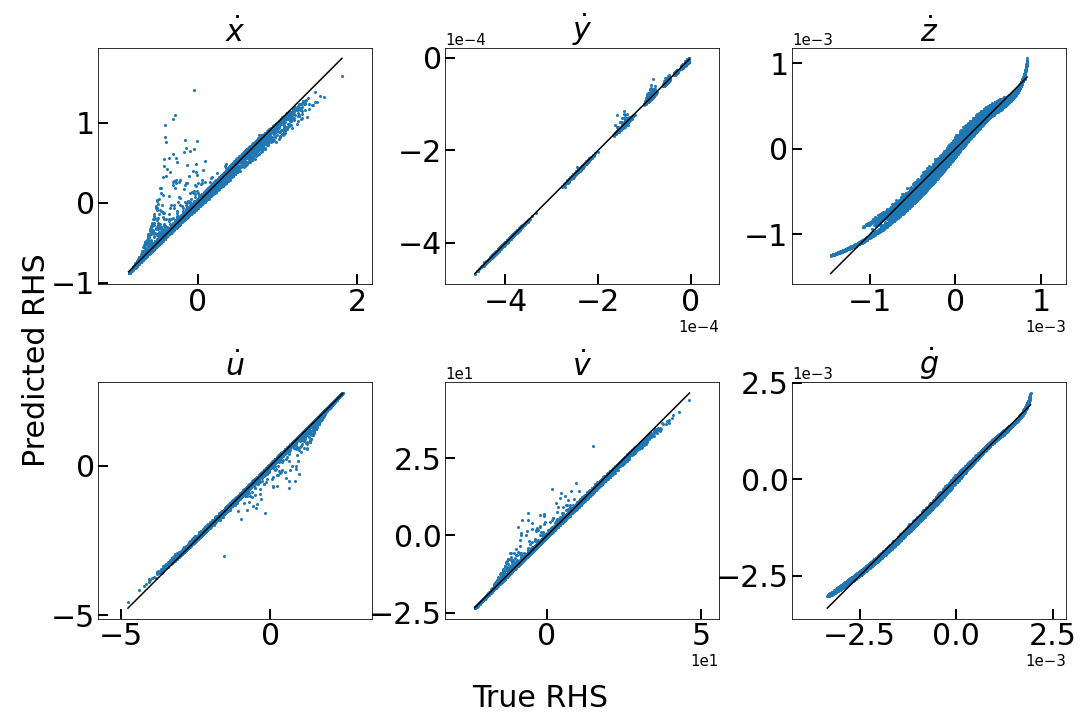}
      \caption{RHS Prediction}
    \end{subfigure}
    \begin{subfigure}[b]{\textwidth}
      \centering
      \includegraphics[width=\textwidth]{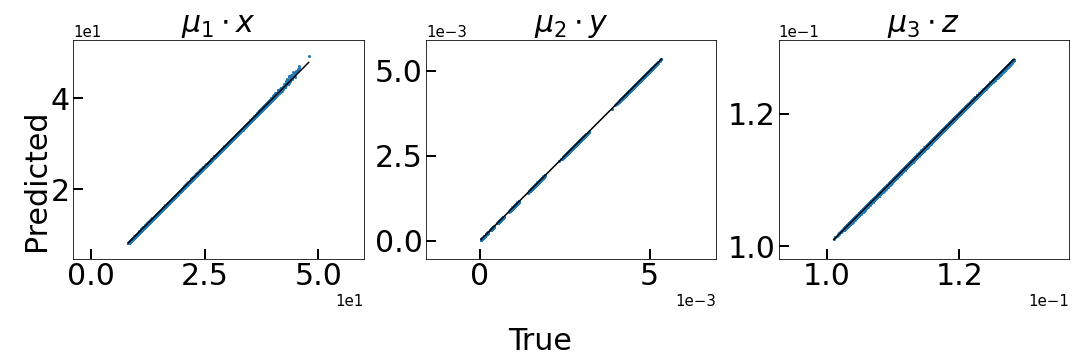}
      \caption{Microbial Growth Function Prediction}
    \begin{subfigure}[b]{\textwidth}
      \centering
      \includegraphics[width=\textwidth]{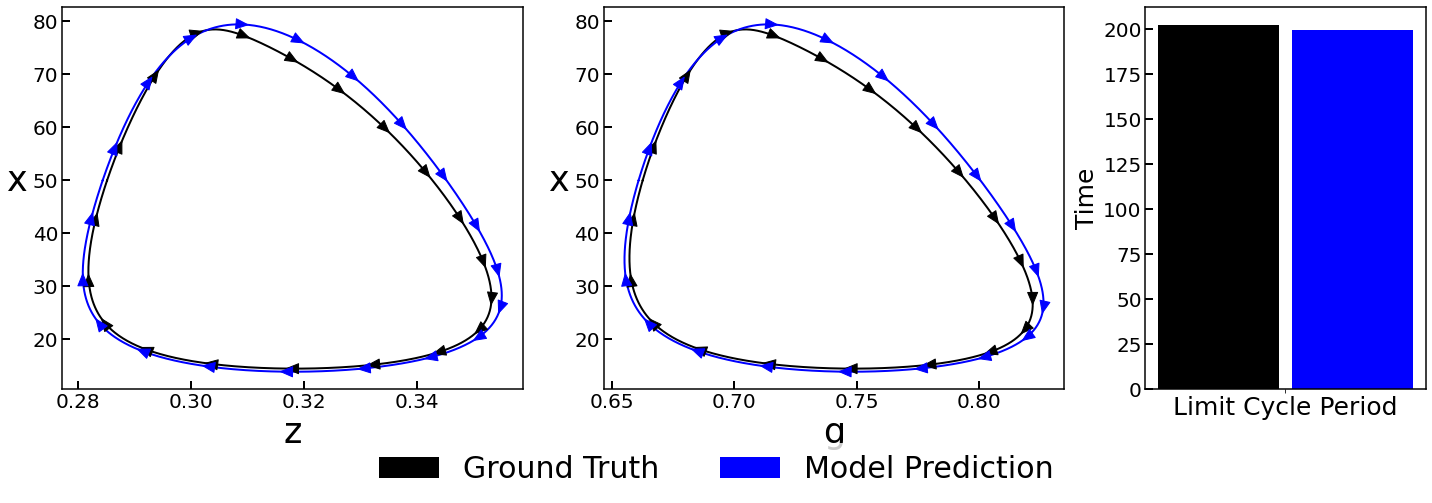}
      \caption{Limit Cycle Prediction}
    \end{subfigure}
    \end{subfigure}
    \end{minipage}
    \hfill
    \begin{minipage}[b]{0.45\textwidth}
    \begin{subfigure}[b]{\textwidth}
      \centering
      \includegraphics[width=\textwidth]{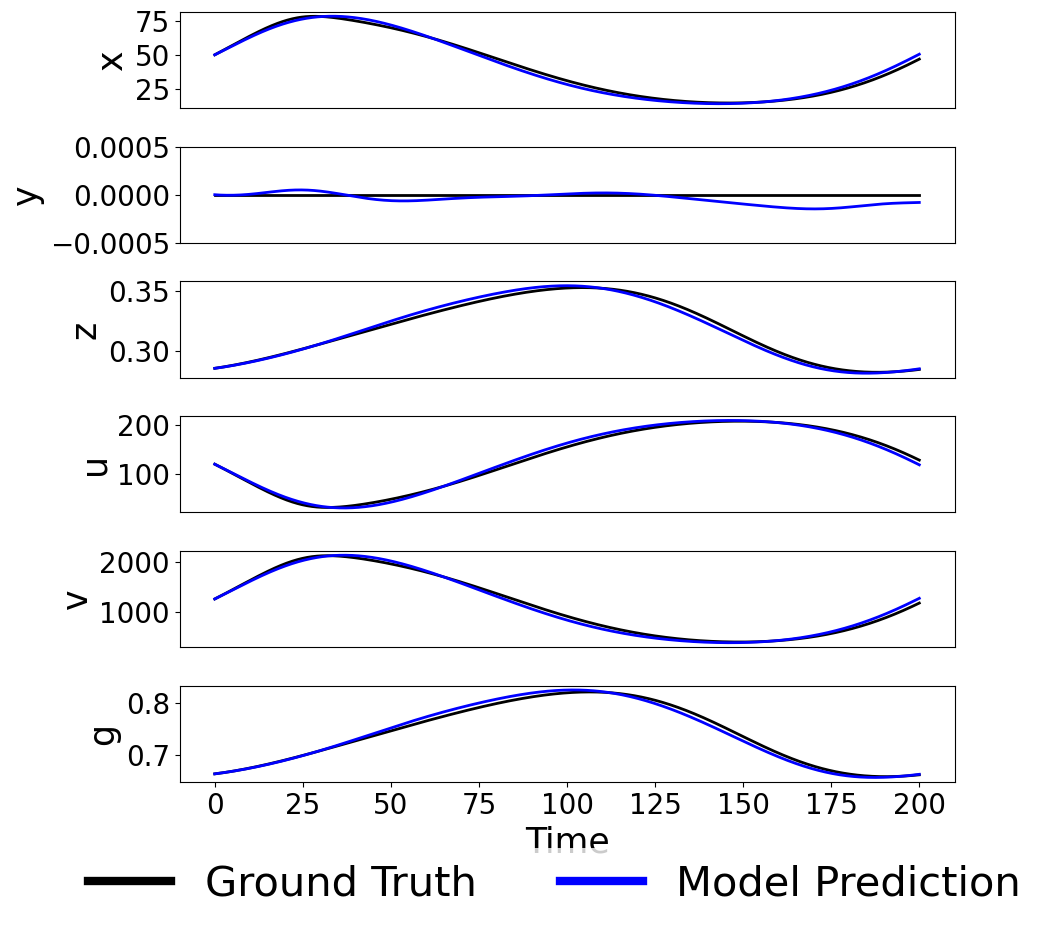}
      \caption{Short Transient Prediction}
    \end{subfigure}
    \begin{subfigure}[b]{\textwidth}
      \centering
      \includegraphics[width=\textwidth]{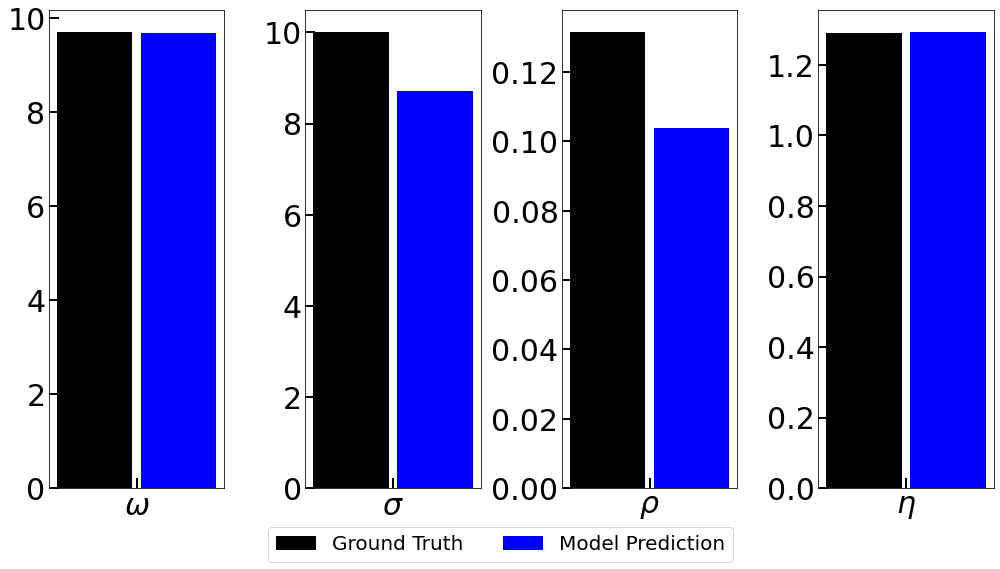}
      \caption{Experimental Parameter Prediction}
    \end{subfigure}
    \end{minipage}

    \caption{Trained model performance for Case BF\_GB2}
    \label{fig:BF_GB2}
\end{figure}

The neural network architecture is a multi-layer perceptron with (a) a six neuron input layer with SiLU activation, (b) two hidden layers of 32 neurons each with SiLU activation, and (c) an output layer with six neurons (black-box formulation) or three neurons (gray-box formulation) with linear activation.
In the ``Black-Box'' formulation, no \textit{a priori} knowledge of the system is assumed, and so the neural network must learn the \textit{full} system (Eq \ref{eq:BF_BB})
\begin{equation}
\begin{aligned}
    \hat{\dot{\mathbf{x}}} &= \mathcal{N}(\mathbf{x};\theta)&&\text{where } \mathbf{x}=\{x,y,z,u,v,g\}.
\end{aligned}
\label{eq:BF_BB}
\end{equation}
In the gray-box formulation, some parts of the system dynamics are known \textit{a priori}; others are unknown (Eq \ref{eq:BF_GB}). Here, the microbial growth rates are assumed to be unknown, but the overall mass and energy conservation laws, as well as the coupling of growth rates to substrate and growth/inhibtion factor production and utilization are known:
\begin{subequations}
\begin{equation}
\begin{aligned}
    \{\hat{\mu}_{1,x}, \hat{\mu}_{2,y}, \hat{\mu}_{3,z}\} &= \mathcal{N}(\mathbf{x};\theta)\\
    \hat{\dot{x}} &= - \alpha x + \hat{\mu}_{1,x} - \mu_{c_1} x\\
    \hat{\dot{y}} &= - \alpha y + \hat{\mu}_{2,y} - \mu_{c_2} y\\
    \hat{\dot{z}} &= - \alpha z + \hat{\mu}_{3,z} - \mu_{c_3} z\\
    \hat{\dot{u}} &= \alpha(u_f-u) - \hat{\mu}_{1,x} \\
    \hat{\dot{v}} &= -\alpha v + \textcolor{blue}{\omega} \hat{\mu}_{1,x} - \hat{\mu}_{2,y} - \textcolor{blue}{\sigma} \hat{\mu}_{3,z}\\
    \hat{\dot{g}} &= -\alpha g + \textcolor{blue}{\rho} \hat{\mu}_{2,y} + \textcolor{blue}{\eta} \hat{\mu}_{3,z}
\end{aligned}
\end{equation}
where $\{\hat{\mu}_{1,x}, \hat{\mu}_{2,y}, \hat{\mu}_{3,z}\}$ are neural network approximations of the microbial growth rates (Eq \ref{eq:BF_true_phi}) multiplied by the biomass concentration of the respective microbe;
and $\{\omega,\sigma,\rho,\eta\}$ are experimental parameters that are either known \textit{a priori}, or else included as \textit{additional} training parameters to be fit during ANN training.
For simplicity of expression, unless otherwise specified, let $\mathcal{N}$ represent the overall network to predict $\hat{\dot{\mathbf{x}}}$, with all gray-box aspects subsumed within this representation.
\begin{equation}
    \hat{\dot{\mathbf{x}}} = \mathcal{N}(\mathbf{x};\theta)
\end{equation}
\label{eq:BF_GB}
\end{subequations}
\begin{table}[h!]
\centering
\resizebox{0.8\textwidth}{!}{%
\begin{tabular}{ |>{\columncolor[gray]{0.8}}c|cccccc| } 
 \hline
 \textbf{Case} & \makecell{Learnable \\ Kinetic \\Functions} & \makecell{Learnable \\ Exp. \\Parameters} & \makecell{ Solution \\ Error From \\True LC ($\mathcal{L}_2$)} & \makecell{ RHS \\ Error ($\mathcal{L}_2$)} & \makecell{Kinetic \\Function \\ Error ($\mathcal{L}_2$)}  & \makecell{Exp. \\Parameter \\ Error ($\mathcal{L}_2$)} \\ 
 \hline
 \textbf{BF\textunderscore BB} & \xmark & \xmark &
$(3.25\pm 0.43)\times 10^{-3}$ 
 &
$(2.12\pm 0.45)\times 10^{-3}$
 & - & - \\
\textbf{ BF\textunderscore GB1} & \cmark & \xmark &
$(5.16\pm 0.68)\times 10^{-3}$
 &
  $(5.43\pm 0.80)\times 10^{-3}$
&
$(7.66\pm 1.14)\times 10^{-4}$
& - \\ 
\textbf{ BF\textunderscore GB2} & \cmark & \cmark & 
$(6.10\pm 2.72)\times 10^{-3}$
 &
$(7.50\pm 3.2)\times 10^{-3}$
&
$(1.09\pm 0.49)\times 10^{-3}$
& 
$(7.51 \pm 2.1) \times 10^{-2}$
\\
\hline
\end{tabular}}
    \caption{Metrics Summary Table for B\&F data. This table showcases model performance of Black-box and Gray-box models. All models have arbitrary data $\Delta t$, partial observations and no initial condition given. Refer Section \ref{sec:Metrics} for details on metrics.}
\label{tab:B&F}
\end{table}
We now demonstrate the training methodology, developed and illustrated through the URP CSTR model, on the B\&F model; the metrics are summarized in Table \ref{tab:B&F}. All cases have the data pathology of arbitrary time sampling, partial observations and with no initial conditions given. Case BF\_BB (Fig \ref{fig:BF_BB}) is the black-box case, and we observe good prediction of the RHS of all 6 variables as well as accurate reproduction of the steady limit cycle.
In cases BF\_GB1 and BF\_GB2 (Fig \ref{fig:BF_GB2}) we introduce physics-informed gray-boxes, with the model learning the microbial growth functions for the species $x,y,z$, with the overall evolution of all 6 variables specified using mass balances with coupling parameters $\omega,\sigma,\rho,\eta$. In Case BF\_GB1 these coupling parameters are known, while in Case BF\_GB2 they are unknown and are introduced as additional parameters to be learned during model training. For Case BF\_GB2 (Fig \ref{fig:BF_GB2}), the model predicts accurately the microbial growth rates and the coupling parameters; this translates to accurate prediction of the RHS of all 6 variables and good reproduction of the limit cycle.

\section{Discussion and Conclusions}

In the course of this work, we observed certain pathologies with the model training that we belive are interesting enough to report and discuss.\\

\begin{figure}[h!]
    \centering
    \begin{subfigure}[b]{0.23\textwidth}
      \centering
      \includegraphics[width=\textwidth]{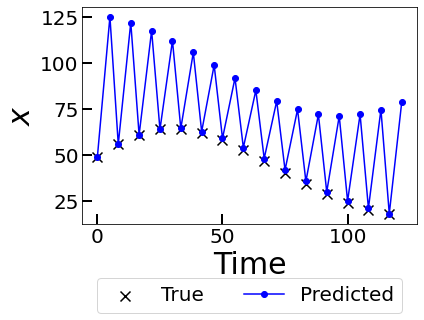}
      \caption{BF\_BB\_A \\ NN Iteration}
    \end{subfigure}
    \begin{subfigure}[b]{0.23\textwidth}
      \centering
      \includegraphics[width=\textwidth]{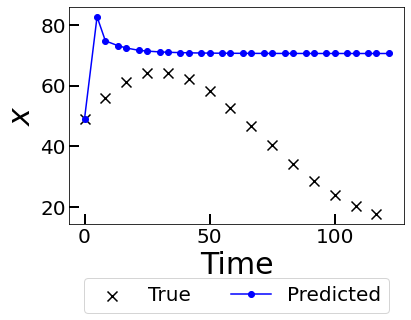}
      \caption{BF\_BB\_A \\ solve\_ivp}
    \end{subfigure}
    \begin{subfigure}[b]{0.23\textwidth}
      \centering
      \includegraphics[width=\textwidth]{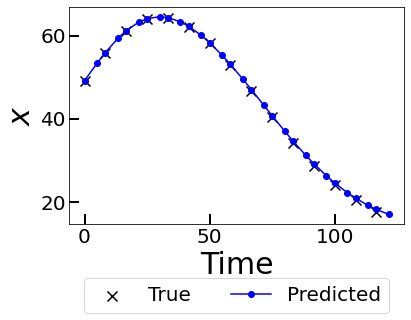}
      \caption{BF\_BB\_B \\ NN Iteration}
    \end{subfigure}
    \begin{subfigure}[b]{0.23\textwidth}
      \centering
      \includegraphics[width=\textwidth]{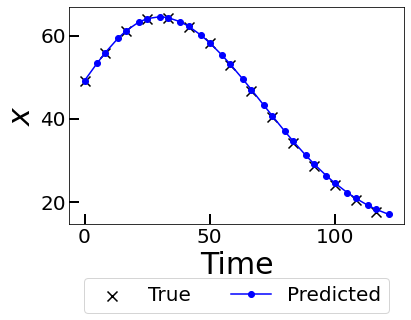}
      \caption{BF\_BB\_B \\ solve\_ivp}
    \end{subfigure}
    \begin{subfigure}[b]{0.23\textwidth}
      \centering
      \includegraphics[width=\textwidth]{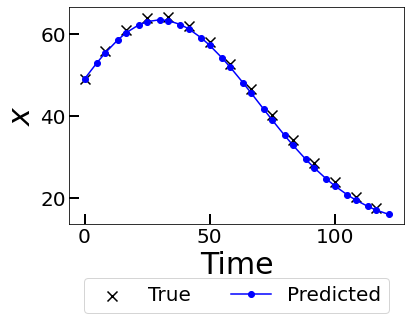}
      \caption{BF\_BB\_C \\ NN Iteration}
    \end{subfigure}
    \begin{subfigure}[b]{0.23\textwidth}
      \centering
      \includegraphics[width=\textwidth]{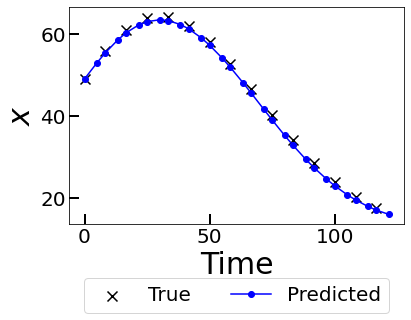}
      \caption{BF\_BB\_C \\ solve\_ivp}
    \end{subfigure}
    \begin{subfigure}[b]{0.23\textwidth}
      \centering
      \includegraphics[width=\textwidth]{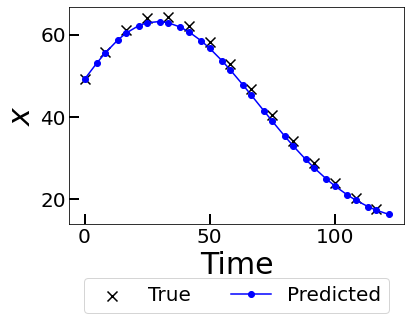}
      \caption{BF\_BB\_D \\ NN Iteration}
    \end{subfigure}
    \begin{subfigure}[b]{0.23\textwidth}
      \centering
      \includegraphics[width=\textwidth]{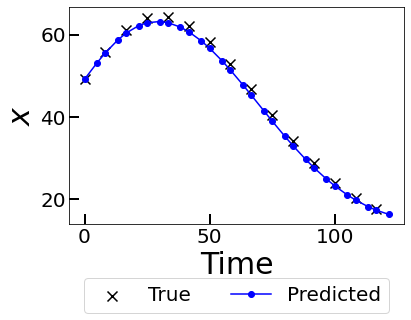}
      \caption{BF\_BB\_D \\ solve\_ivp}
    \end{subfigure}
    \caption{The performance of the best model from each case on a single representative training trajectory. This is shown
    for both (a) the recursive neural network iteration on the 
    with the time-steps used in the BF\_BB\_A training; and (b)  
    the scipy.integrate.solve\_ivp solver, which uses variable time stepping (and thus helps better test the accuracy of integrating the learned  RHS).
    }
    \label{fig:BF_resonance}
\end{figure}

\noindent{\bf The ``Resonance'' Effect.}  We observed that, when working with fixed rate time sampling data, depending on the initialization of model weights, training might indeed successfully fit the data by converging to a ``visibly wrong" local minimum. 
Rather than learning the RHS of the underlying ODE system,  it learned a ``resonance-type" time-one map: instead of a smooth curve interpolating the data points, two integration steps, with very different slopes (Fig \ref{fig:BF_resonance} (a),(b)), now take us from each data point to the next; one can describe this as a 2:1 resonance between the integration step and the data sampling step. 

In the above example, the data was sampled every $\Delta t=8.333$. For case Bf\_BB\_A, this $\Delta t$ was broken up in the RK4 numerical integrator into steps of $(5+3.333)$ consistently. With ``bad" (random) weight initialization, the model training ``hard-wires'' this time step, essentially learning a time-one map without explicitly learning the RHS of the ODE; when iterated through the network with the same time-steps as during training, we observe a characteristic zig-zag ``resonance'' style output.
We explored various mitigation techniques to prevent convergence to such a local minimum. One approach involved attempts at better initialization of the model weights (here, simply scaling the output layer's weights down by a factor of 100 (Case BF\_BB\_B), with the initialization of other layers unchanged). Secondly, we randomize the solver $\delta t$ steps so that, rather than consistently taking steps of $(5+3.333)$, the $\Delta t=8.333$ was broken up into randomly sized time ``chunks", all selected below the max $\delta t$ of $5$ (Case BF\_BB\_C). Thirdly, we randomize the data sampling itself, with a gamma distribution with mean $<\Delta t=8.333>$, and in doing so the solver $\delta t$ values were naturally randomized too. All three methods allowed the model to escape the bad local minimum, or to even remove it entirely.

\begin{table}[h!]
\centering
\resizebox{0.8\textwidth}{!}{%
\begin{tabular}{ |>{\columncolor[gray]{0.8}}c|ccccc| } 
 \hline
 \textbf{Case} & \makecell{Initial \\ Scaling} & \makecell{Randomized \\ Data \\ $\Delta t$} & \makecell{Randomized \\ Solver \\ $\delta t$}& \makecell{ Solution \\ Error From \\True LC ($\mathcal{L}_2$)} & \makecell{ RHS \\ Error ($\mathcal{L}_2$)}\\
 \hline
\textbf{ BF\textunderscore BB\_A} & $\times 100^{0}$ & \xmark & \xmark &
$(4.33\pm 2.58)\times 10^{-1}$
&
$8.47\pm 1.63$
 \\
 \hline 
\textbf{ BF\textunderscore BB\_B} & $\times 100^{-1}$ & \xmark & \xmark &
$(4.13\pm 0.53)\times 10^{-3}$ 
&
$(2.89\pm 0.92)\times 10^{-3}$
\\
 \hline
  \textbf{BF\textunderscore BB\_C} & $\times 100^{0}$ & \xmark & \cmark &
$(8.40\pm 3.52)\times 10^{-2}$
&
$1.65\pm 0.68$
\\
 \hline
  \textbf{BF\textunderscore BB\_D} & $\times 100^{0}$ & \cmark & \makecell{\cmark \\ (Because \\ of data)} &
$(1.14\pm 0.17)\times 10^{-2}$
&
$(2.88\pm 0.68)\times 10^{-2}$
\\
 \hline
  \textbf{BF\textunderscore BB\_E} & $\times 100^{-1}$ & \cmark & \makecell{\cmark \\ (Because \\ of data)} &
$(4.18\pm 0.71)\times 10^{-3}$
&
$(3.10\pm 0.56)\times 10^{-3}$
\\
 \hline
\end{tabular}}
    \caption{Metrics Summary Table for B\&F data. This table showcases the resonance effect. Refer Section \ref{sec:Metrics} for details on metrics.}
\end{table}

\begin{figure}[h!]
    \begin{subfigure}[b]{0.28\textwidth}
      \centering
      \includegraphics[width=.95\textwidth]{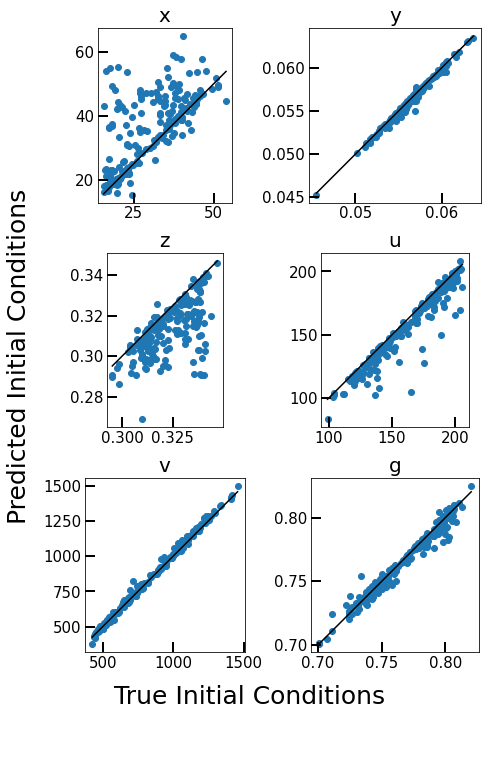}
      \caption{Initial Conditions \\Prediction}
    \end{subfigure}
    \begin{subfigure}[b]{0.26\textwidth}
      \centering
      \includegraphics[width=.95\textwidth]{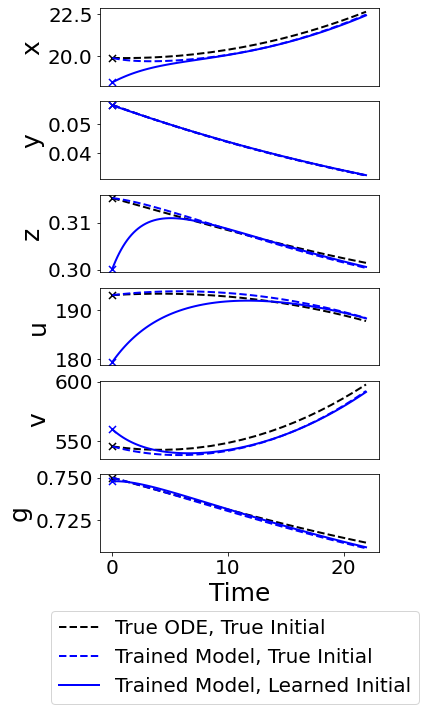}
      \caption{Transient \\(Zoomed)}
    \end{subfigure}
    \begin{subfigure}[b]{0.44\textwidth}
      \centering
      \includegraphics[width=.95\textwidth]{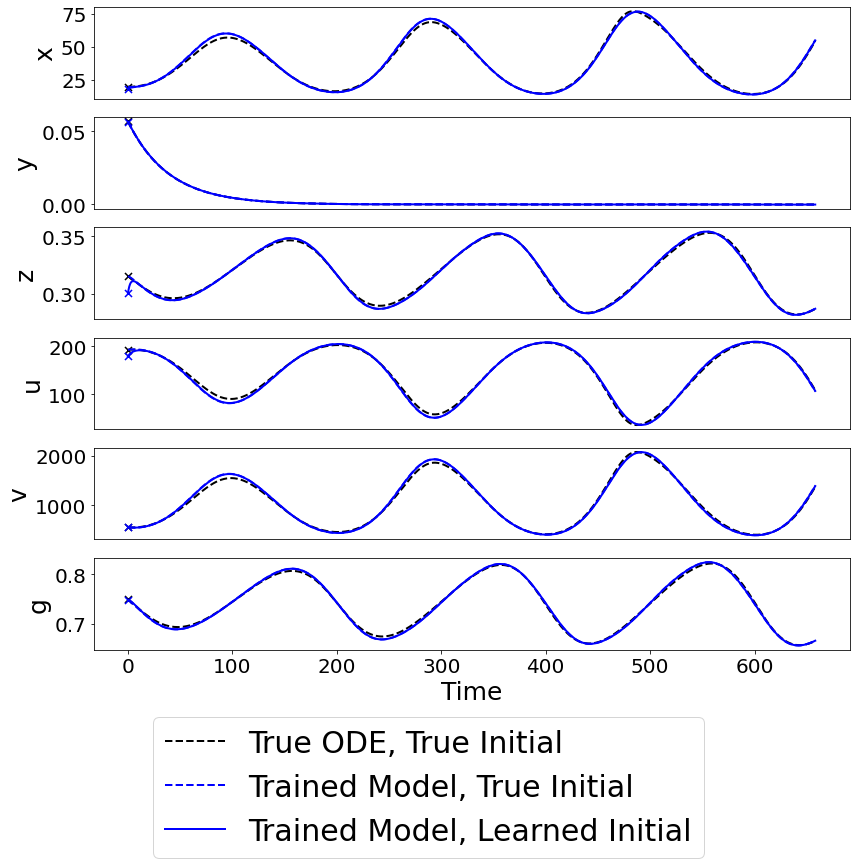}
      \caption{Transient}
    \end{subfigure}
    \caption{Initial Condition Prediction for Case BF\_GB2. Despite the initial conditions learned not being correct, the \textit{trained} ANN when integrated from the learned initial condition (solid blue line) quickly converges with the trajectory integrated from the true initial condition (dashed blue line), and in long-term dynamics the two are practically indistinguishable. This is due to the separation of time scales (fast-slow nature) of the initial startup.}
    \label{fig:BF_IC}
\end{figure}

\begin{figure}[h!]
    \centering
        \begin{equation}
            \begin{aligned}
                \dot{x} &= f(x,y)\\
                \varepsilon\dot{y}&=g(x,y)
            \end{aligned}
            \label{eq:singularperturb}
        \end{equation}
      \centering
      \includegraphics[width=.5\textwidth]{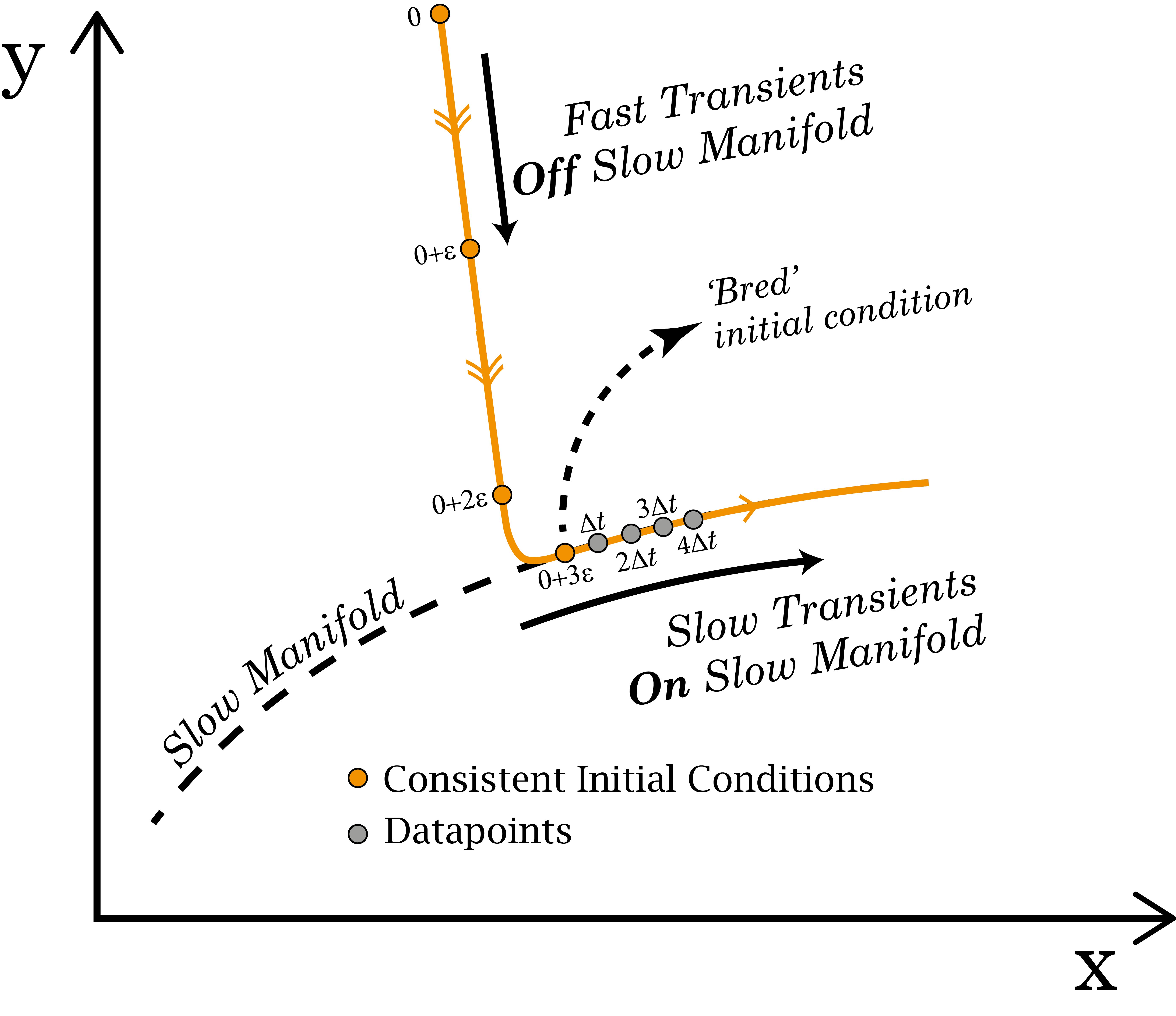}
      \caption{Separation of fast and slow time-scales, as observed in singular perturbation problems (Eq \ref{eq:singularperturb}), can lead to multiple initial conditions that are \textit{practically} consistent with the same first measured datapoint, as they evolve rapidly on the fast manifold before converging on to the slow manifold. Ideally, we would want the initial condition that has been `bred' till it lies on the slow manifold as this is the ``most likely'' candidate, but this can be hard to impose.}
      \label{fig:singularperturb}
\end{figure}

\noindent{\bf The learning of initial conditions.}
We have observed that when initial conditions were not part of the training set, and needed to be inferred our models did \textbf{not} consistently infer the correct initial conditions (Fig \ref{fig:BF_IC}).
This can be rationalized (and mitigated) by considering what happens in multi-time-scale singularly perturbed dynamic problems (fast-slow systems, systems with large separation of time scales) (Fig \ref{fig:singularperturb}, Eq \ref{eq:singularperturb}).
In such problems (and our 6 ODEs exhibit eigenvalues of the linearization that vary by one or more orders of magnitude (Table \ref{tab:BF_res_eig})), 
initial conditions very quickly (over the fast time scale) converge to a
``slow manifold", on which the long-term dynamics evolve over the  slow time scale.
This is the same argument as the celebrated Quasi-Steady-State (QSSA) or Bodenstein Approximation in chemical kinetics.
This means that {\em an infinity} of initial conditions will quickly (over the fast time scale) end up at (practically) the same state value on the slow manifold after this short initial integration period, and then remain and evolve on the slow manifold \cite{Gear2005, Vandekerckhove2009, Antonios2012, Zagaris2009}.
%
%
%

In the relevant literature \cite{Balmaseda2009}, this is mitigated by searching
for an initial condition that {\em already} lies approximately on the slow manifold:
Say we are interested in an initial condition $\Delta t$ before our first sampled data point.
The ``trick" for constructing such a ``mature" or ``bred" initial condition 
%
one $\Delta t$ in the past involves estimating an initial condition farther back (say 10 $\Delta t$ earlier, which can be anywhere on the fast foliation of the slow manifold), and then let it evolve for $9 \Delta t$.  This allows the ``bad" components of the earlier initial condition to ``mature" (``breed") giving us a good estimate for an initial condition {\em on the slow manifold} at the desired one $\Delta t$ in the past. 
\begin{table}[h!]
\centering
\resizebox{0.8\textwidth}{!}{%
\begin{tabular}{ |c|c|c|c| } 
 \hline
 \textbf{Eigenvalue} & \makecell{\textbf{True ODE} \\ \textbf{True IC}} & \makecell{\textbf{Trained ANN} \\ \textbf{True IC}} & \makecell{\textbf{Trained ANN} \\ \textbf{Learned IC}} \\
 \hline
 1 & $-0.20506 -0.020221j$ & \textcolor{red}{$-0.51397$} & \textcolor{red}{$-0.51752$} \\
 2 & $-0.20506 +0.020221j$ & \textcolor{red}{$-0.44970$} & \textcolor{red}{$-0.44258$} \\
 3 & \textcolor{blue}{$-0.20145$} & \textcolor{blue}{$-0.20145$} & \textcolor{blue}{$-0.20145$} \\
 4 & \textcolor{blue}{$-0.025494$} & \textcolor{blue}{$-0.026535$} & \textcolor{blue}{$-0.026341$}\\
 5 & $0.002033-0.03847j$ & $-0.000690-0.03661j$ & $-0.004094-0.03534j$ \\
 6 & $0.002033+0.03847j$ & $-0.000690+0.03661j$ & $-0.004094+0.03534j$ \\
 \hline
\end{tabular}}
    \caption{Eigenvalues for the jacobian of the ground truth ODEs and the trained model on the true and learned initial conditions. The fast dynamics associated with the leading eigenvalues (in red) decay quickly at a time scale of $t=2-2.3$ and quickly converge to the slower manifolds associated with the eigenvalues in blue that are consistent with the true system. The training data used was sampled at an average $\Delta t=8$, hence the faster dynamics could not be learned (or discouraged) during model training.}
    \label{tab:BF_res_eig}
\end{table}
%

\noindent{\bf Summary and Outlook.}
In this paper, we presented a neural-network based identification architecture---inspired by, and based on traditional numerical analysis---to address a number of time series data pathologies: uneven sampling rates, missing data (including initial conditions), as well unknown or partially known physics. 

We formulated the systems identification process as the learning of the right-hand-side of a system of ordinary differential equations, which can be combined with numerical integration methods.
In each forward pass, we iterate through each training trajectory through time, using teacher-forcing when real data is available, and autoregressive iterations when not, allowing for training on partial observations. 
We demonstrated how this architecture can be integrated with white-box prior knowledge, where the neural networks learn physically interpretable functions such as microbial growth rates or chemical reaction rates, as well as experimental parameters; global laws, such as coupling between reaction rates and heats of reaction, or microbial growth rates, substrate consumption and product formation were easy to systematically ``hardwire" in the architecture. 

We expect that the approach (especially its ``gray box" component) will make it useful in industrial applications (e.g. in CHO cell culture biomanufacturing, which we are currently exploring).
On the technical/computational side, a natural next task involves the efficient and easily usable implementation of parallel batching, and, more generally the consistent use of parallelization to accelerate computation. 
We also expect work towards incorporating adaptivity of the approach, in the spirit of adaptive time step selection for traditional initial value solvers.
Linking this work with Uncertainty Quantification is an important direction, and a subject of current applied ML research, whether on parameter uncertainty or on prediction uncertainty. 
Remarkably, the error between the identified right-hand-sides of the dynamics (termed {\em Inverse Modified Differential Equations} or IMDEs) and the true dynamics---the so called  {\em  Inverse Modified Error Analysis}---appears to be a nascent branch of numerical analysis, born by precisely the type of work we present here \cite{Zhu2022, Zhu2023}\\

\noindent{\bf Author Contributions.} IGK planned research, supervised research, and wrote manuscript. SM, TB and TQ performed research and wrote manuscript. JLA and MB contributed to modeling issues and writing manuscript.

\noindent{\bf Funding Acknowledgements.}
The work of IGK and TB was partially supported by the US Department of Energy (DOE) (grant number SA22-0052-S001); and SM and JLA  were supported by the DOE (grant number DE-SC0019363). IGK and TB were also partially supported by the US Air Force Office of Scientific Research (AFOSR) (grant number FA9550-21-0317); and by AMBIC, that partially supported the work of MB. SM was also supported by the National Science Scholarship (NSS) from the Agency for Science, Technology and Research (A*STAR) Singapore.\\

\noindent{\bf Competing Interests.} The authors declare no competing interests.


\clearpage

\bibliographystyle{plain}
\bibliography{references}

\appendix
\setcounter{table}{0}
\renewcommand{\thetable}{\thesection\arabic{table}}
\setcounter{equation}{0}
\renewcommand{\theequation}{\thesection\arabic{equation}}
\setcounter{figure}{0}
\renewcommand{\thefigure}{\thesection\arabic{figure}}

\section{Supplementary}

\subsection{Supplementary Results}

\subsubsection{URP Model}

\begin{table}[H]
\centering
\resizebox{0.8\textwidth}{!}{%
\begin{tabular}{ |c|c|c|c|c|c|c|c| } 
 \hline
 Case & \makecell{Data \\ $<\Delta t>$} & \makecell{Solver \\ max \\ $\delta t$} & \makecell{ Arbitrary \\ Data $\Delta t$} & \makecell{ Partial \\ Observations} & \makecell{ Initial \\ Condition \\ Given} & \makecell{ Solution \\ Error} & \makecell{ RHS \\ Error} 
\\ 
 \hline
 A & 0.1 & 0.1 & \xmark & \xmark & \cmark &
   \makecell{$\mathcal{L}_1: (1.31\pm 0.24)\times 10^{-2}$ \\ $\mathcal{L}_2: (9.68\pm 1.71)\times 10^{-3}$ \\ $\mathcal{L}_\infty: (1.58\pm 0.27)\times 10^{-2}$} 
 & 
  \makecell{$\mathcal{L}_1: (3.66\pm 0.52)\times 10^{-3}$ \\ $\mathcal{L}_2: (2.72\pm 0.36)\times 10^{-3}$ \\ $\mathcal{L}_\infty: (4.56\pm 0.57)\times 10^{-3}$} 
 \\\hline
 B & 0.5 & 0.5 & \xmark & \xmark & \cmark &
    \makecell{$\mathcal{L}_1: (1.97\pm 0.17)\times 10^{-2}$ \\ $\mathcal{L}_2: (1.49\pm 0.12)\times 10^{-2}$ \\ $\mathcal{L}_\infty: (2.58\pm 0.22)\times 10^{-2}$} 
 & 
     \makecell{$\mathcal{L}_1: (1.32\pm 0.10)\times 10^{-2}$ \\ $\mathcal{L}_2: (9.71\pm 0.76)\times 10^{-3}$ \\ $\mathcal{L}_\infty: (1.61\pm 0.13)\times 10^{-2}$} 
 \\ \hline
 C & 0.5 & 0.1 & \xmark & \xmark & \cmark &
  \makecell{$\mathcal{L}_1: (1.14\pm 0.15)\times 10^{-2}$ \\ $\mathcal{L}_2: (8.69\pm 1.18)\times 10^{-3}$ \\ $\mathcal{L}_\infty: (1.50\pm 0.24)\times 10^{-2}$} 
 & 
\makecell{$\mathcal{L}_1: (7.90\pm 1.64)\times 10^{-3}$ \\ $\mathcal{L}_2: (5.95\pm 1.21)\times 10^{-3}$ \\ $\mathcal{L}_\infty: (1.03\pm 0.21)\times 10^{-2}$} 
 \\ \hline
 D & 0.5 & 0.1 & \cmark & \xmark & \cmark &
 \makecell{$\mathcal{L}_1: (5.56\pm 0.60)\times 10^{-3}$ \\ $\mathcal{L}_2: (4.23\pm 0.43)\times 10^{-3}$ \\ $\mathcal{L}_\infty: (7.44\pm 0.80)\times 10^{-3}$} 
 & 
 \makecell{$\mathcal{L}_1: (4.67\pm 0.44)\times 10^{-3}$ \\ $\mathcal{L}_2: (3.50\pm 0.34)\times 10^{-3}$ \\ $\mathcal{L}_\infty: (5.97\pm 0.62)\times 10^{-3}$} 
 \\ \hline
 E & \makecell{0.5 ($x_1$)\\0.55 ($x_2$)} & 0.1 & \xmark & \cmark & \cmark &
  \makecell{$\mathcal{L}_1: (8.08\pm 0.79)\times 10^{-3}$ \\ $\mathcal{L}_2: (6.09\pm 0.56)\times 10^{-3}$ \\ $\mathcal{L}_\infty: (1.04\pm 0.09)\times 10^{-2}$}
 & 
   \makecell{$\mathcal{L}_1: (4.91\pm 0.63)\times 10^{-3}$ \\ $\mathcal{L}_2: (3.66\pm 0.46)\times 10^{-3}$ \\ $\mathcal{L}_\infty: (6.27\pm 0.75)\times 10^{-3}$}
 \\ \hline
 F & 0.5 & 0.1 & \cmark & \cmark & \cmark &
\makecell{$\mathcal{L}_1: (1.16\pm 0.22)\times 10^{-2}$ \\ $\mathcal{L}_2: (8.91\pm 1.65)\times 10^{-3}$ \\ $\mathcal{L}_\infty: (1.59\pm 0.30)\times 10^{-2}$}
 & 
\makecell{$\mathcal{L}_1: (9.22\pm 1.05)\times 10^{-3}$ \\ $\mathcal{L}_2: (7.12\pm 0.81)\times 10^{-3}$ \\ $\mathcal{L}_\infty: (1.27\pm 0.15)\times 10^{-2}$}
 \\ \hline
 URP\_BB & 0.5 & 0.1 & \cmark & \cmark & \xmark &
 \makecell{$\mathcal{L}_1: (9.91\pm 1.71)\times 10^{-3}$ \\ $\mathcal{L}_2: (7.92\pm 1.36)\times 10^{-3}$ \\ $\mathcal{L}_\infty: (1.45\pm 0.26)\times 10^{-2}$}
 & 
\makecell{$\mathcal{L}_1: (9.99\pm 1.41)\times 10^{-3}$ \\ $\mathcal{L}_2: (7.69\pm 0.11)\times 10^{-3}$ \\ $\mathcal{L}_\infty: (1.37\pm 0.18)\times 10^{-2}$}
 \\ 
 \hline
\end{tabular}}
\caption{Metrics Summary Table for URP CSTR data. This table showcases model performance on data of increasing complexity of representation. Refer Section \ref{sec:Metrics} for details on metrics.}
\label{sup:tab:UPR_Metrics_Cases}
\end{table}

\begin{table}[H]
\centering
\resizebox{0.8\textwidth}{!}{%
\begin{tabular}{ |c|c|c|c|c|c|c| } 
 \hline
 Case & \makecell{Learnable \\ Kinetic \\Functions} & \makecell{Learnable \\ Experimental \\Parameters} & \makecell{ Solution \\ Error} & \makecell{ RHS \\ Error}  & \makecell{ Kinetic \\Function \\ Error} & \makecell{ Experimental \\Parameter \\ Error}\\ 
 \hline
 URP\textunderscore BB& \xmark & \xmark &
 \makecell{$\mathcal{L}_1: (9.91\pm 1.71)\times 10^{-3}$ \\ $\mathcal{L}_2: (7.92\pm 1.36)\times 10^{-3}$ \\ $\mathcal{L}_\infty: (1.45\pm 0.26)\times 10^{-2}$}
 & 
\makecell{$\mathcal{L}_1: (9.99\pm 1.41)\times 10^{-3}$ \\ $\mathcal{L}_2: (7.69\pm 0.11)\times 10^{-3}$ \\ $\mathcal{L}_\infty: (1.37\pm 0.18)\times 10^{-2}$}
 &-&-\\ \hline
 URP\textunderscore GB1& \cmark & \xmark &
\makecell{$\mathcal{L}_1: (1.37\pm 0.18)\times 10^{-2}$ \\ $\mathcal{L}_2: (9.91\pm 1.26)\times 10^{-3}$ \\ $\mathcal{L}_\infty: (1.53\pm 0.18)\times 10^{-2}$}
 & 
\makecell{$\mathcal{L}_1: (6.24\pm 0.72)\times 10^{-3}$ \\ $\mathcal{L}_2: (4.43\pm 0.51)\times 10^{-3}$ \\ $\mathcal{L}_\infty: (6.87\pm 0.80)\times 10^{-3}$}
 &
$(5.31\pm0.62) \times 10^{-3}$
 &-\\ \hline
 URP\textunderscore GB2& \cmark & \cmark &
\makecell{$\mathcal{L}_1: (1.68\pm 0.24)\times 10^{-2}$ \\ $\mathcal{L}_2: (1.22\pm 0.17)\times 10^{-2}$ \\ $\mathcal{L}_\infty: (1.90\pm 0.26)\times 10^{-2}$}
 & 
\makecell{$\mathcal{L}_1: (8.97\pm 2.87)\times 10^{-3}$ \\ $\mathcal{L}_2: (6.49\pm 2.08)\times 10^{-3}$ \\ $\mathcal{L}_\infty: (1.04\pm 0.34)\times 10^{-2}$}
 &
$(7.77\pm 2.25) \times 10^{-3}$
 &
 $(4.45\pm 2.28)\times 10^{-2}$
 \\ 
 \hline
\end{tabular}}

\caption{Metrics Summary Table for URP CSTR data. This table showcases model performance of Black-box and Gray-box models. All models have arbitrary data $\Delta t$, partial observations and no initial condition given. Refer Section \ref{sec:Metrics} for details on metrics.}
\label{sup:tab:UPR_Metrics_BBGB}
\end{table}

\subsubsection{B\&F Model}

\begin{table}[H]
\centering
\resizebox{1\textwidth}{!}{%
\begin{tabular}{ |c|c|c|c|c|c|c| } 
 \hline
 Case & \makecell{Learnable \\ Kinetic \\Functions} & \makecell{Learnable \\ Exp. \\Parameters} & \makecell{ Solution \\ Error \\From \\True LC} & \makecell{ RHS \\ Error} & \makecell{Kinetic \\Function \\ Error}  & \makecell{Exp. \\Parameter \\ Error} \\ 
 \hline
 BF\textunderscore BB & \xmark & \xmark &
\makecell{$\mathcal{L}_1: (6.26\pm 0.81)\times 10^{-3}$ \\ $\mathcal{L}_2: (3.25\pm 0.43)\times 10^{-3}$ \\ $\mathcal{L}_\infty: (1.36\pm 0.17)\times 10^{-2}$} 
 &
 \makecell{$\mathcal{L}_1: (4.00\pm 0.77)\times 10^{-3}$ \\ $\mathcal{L}_2: (2.12\pm 0.45)\times 10^{-3}$ \\ $\mathcal{L}_\infty: (7.13\pm 1.28)\times 10^{-3}$} 
 & - & - \\
 \hline 
 BF\textunderscore GB1 & \cmark & \xmark &
 \makecell{$\mathcal{L}_1: (1.04\pm 0.15)\times 10^{-2}$ \\ $\mathcal{L}_2: (5.16\pm 0.68)\times 10^{-3}$ \\ $\mathcal{L}_\infty: (1.90\pm 0.24)\times 10^{-2}$} 
 &
  \makecell{$\mathcal{L}_1: (9.94\pm 1.41)\times 10^{-3}$ \\ $\mathcal{L}_2: (5.43\pm 0.80)\times 10^{-3}$ \\ $\mathcal{L}_\infty: (1.68\pm 0.23)\times 10^{-2}$} 
&
  \makecell{$\mathcal{L}_1: (1.02\pm 0.14)\times 10^{-3}$ \\ $\mathcal{L}_2: (7.66\pm 1.14)\times 10^{-4}$ \\ $\mathcal{L}_\infty: (1.24\pm 0.17)\times 10^{-3}$} 
& - \\ \hline
 BF\textunderscore GB2 & \cmark & \cmark & 
 \makecell{$\mathcal{L}_1: (1.22\pm 0.57)\times 10^{-2}$ \\ $\mathcal{L}_2: (6.10\pm 2.72)\times 10^{-3}$ \\ $\mathcal{L}_\infty: (2.33\pm 1.09)\times 10^{-2}$} &
\makecell{$\mathcal{L}_1: (1.37\pm 0.60)\times 10^{-2}$ \\ $\mathcal{L}_2: (7.50\pm 3.2)\times 10^{-3}$ \\ $\mathcal{L}_\infty: (2.47\pm 1.48)\times 10^{-2}$} 
&
  \makecell{$\mathcal{L}_1: (1.48\pm 0.71)\times 10^{-3}$ \\ $\mathcal{L}_2: (1.09\pm 0.49)\times 10^{-3}$ \\ $\mathcal{L}_\infty: (1.79\pm 1.17)\times 10^{-3}$} 
& 
$(7.51 \pm 2.1) \times 10^{-2}$
\\
\hline
\end{tabular}}
    \caption{Metrics Summary Table for B\&F data. This table showcases model performance of Black-box and Gray-box models. All models have arbitrary data $\Delta t$, partial observations and no initial condition given. Errors are means and standard deviations of 10 training runs. Refer to Section \ref{sec:Metrics} for details on error metrics..}
\label{sup:tab:B&F}
\end{table}

\begin{table}[H]
\centering
\resizebox{0.8\textwidth}{!}{%
\begin{tabular}{ |c|c|c|c|c|c| } 
 \hline
 Case & \makecell{Initial \\ Scaling} & \makecell{Randomized \\ Data \\ $\Delta t$} & \makecell{Randomized \\ Solver \\ $\delta t$}& \makecell{ Solution \\ Error \\From \\True LC} & \makecell{ RHS \\ Error}\\
 \hline
 BF\textunderscore BB\_A & $\times 100^{0}$ & \xmark & \xmark &
 \makecell{$\mathcal{L}_1: (8.02 \pm 4.02)\times 10^{-1}$ \\ $\mathcal{L}_2: (4.33\pm 2.58)\times 10^{-1}$ \\ $\mathcal{L}_\infty: 2.03\pm 1.43$} 
&
 \makecell{$\mathcal{L}_1: 14.6 \pm 2.56$ \\ $\mathcal{L}_2: 8.47\pm 1.63$ \\ $\mathcal{L}_\infty: 38.3\pm 9.43$} 
 \\
 \hline 
 BF\textunderscore BB\_B & $\times 100^{-1}$ & \xmark & \xmark &
 \makecell{$\mathcal{L}_1: (7.84 \pm 0.92)\times 10^{-3}$ \\ $\mathcal{L}_2: (4.13\pm 0.53)\times 10^{-3}$ \\ $\mathcal{L}_\infty: (1.72\pm 0.24) \times 10^{-2}$} 
&
 \makecell{$\mathcal{L}_1: (5.29 \pm 1.50)\times 10^{-3}$ \\ $\mathcal{L}_2: (2.89\pm 0.92)\times 10^{-3}$ \\ $\mathcal{L}_\infty: (9.23\pm 2.49) \times 10^{-3}$} 
\\
 \hline
  BF\textunderscore BB\_C & $\times 100^{0}$ & \xmark & \cmark &
 \makecell{$\mathcal{L}_1: (1.63 \pm 0.70)\times 10^{-1}$ \\ $\mathcal{L}_2: (8.40\pm 3.52)\times 10^{-2}$ \\ $\mathcal{L}_\infty: (3.50\pm 1.43) \times 10^{-1}$} 
&
 \makecell{$\mathcal{L}_1: 2.77\pm 1.22$ \\ $\mathcal{L}_2: 1.65\pm 0.68$ \\ $\mathcal{L}_\infty: 7.59\pm 3.38$} 
\\
 \hline
  BF\textunderscore BB\_D & $\times 100^{0}$ & \cmark & \makecell{\cmark \\ (Because \\ of data)} &
 \makecell{$\mathcal{L}_1: (2.29 \pm 0.39)\times 10^{-2}$ \\ $\mathcal{L}_2: (1.14\pm 0.17)\times 10^{-2}$ \\ $\mathcal{L}_\infty: (4.46\pm 0.66) \times 10^{-2}$} 
&
 \makecell{$\mathcal{L}_1: (4.57 \pm 1.00)\times 10^{-2}$ \\ $\mathcal{L}_2: (2.88\pm 0.68)\times 10^{-2}$ \\ $\mathcal{L}_\infty: (1.21\pm 0.31) \times 10^{-1}$} 
\\
 \hline
  BF\textunderscore BB\_E & $\times 100^{-1}$ & \cmark & \makecell{\cmark \\ (Because \\ of data)} &
 \makecell{$\mathcal{L}_1: (7.99 \pm 1.33)\times 10^{-3}$ \\ $\mathcal{L}_2: (4.18\pm 0.71)\times 10^{-3}$ \\ $\mathcal{L}_\infty: (1.74\pm 0.30) \times 10^{-2}$} 
&
 \makecell{$\mathcal{L}_1: (5.69 \pm 1.01)\times 10^{-3}$ \\ $\mathcal{L}_2: (3.10\pm 0.56)\times 10^{-3}$ \\ $\mathcal{L}_\infty: (1.02\pm 0.18) \times 10^{-2}$} 
\\
 \hline
\end{tabular}}
    \caption{Metrics Summary Table for B\&F data. This table showcases the resonance effect. Refer Section \ref{sec:Metrics} for details on metrics.}
\end{table}

\subsection{Metrics Used} \label{sec:Metrics}
\begin{equation}
\begin{aligned}
    \dot{\mathbf{x}} &= f(\mathbf{x},\mathbf{p}) \;\;\;\;\;\;\;\;\;\;\;\;\;\;\;\;\;\;\;\;\;True\;\;ODE\\
    \hat{\dot{\mathbf{x}}} &=
    \begin{cases}
    \mathcal{N}(\mathbf{x},\mathbf{p};\theta) & Learned\;\;ODE\;\;(Black\;\;Box)\\
    g(\mathbf{x}, \mathbf{p}, \mathcal{N}(\mathbf{x},\mathbf{p};\theta)) & Learned\;\;ODE\;\;(Gray\;\;Box)\\
    \end{cases}
\end{aligned}
\end{equation}

\begin{figure}[h!]
    \centering
    \begin{subfigure}[b]{0.45\textwidth}
        \includegraphics[width=\textwidth]{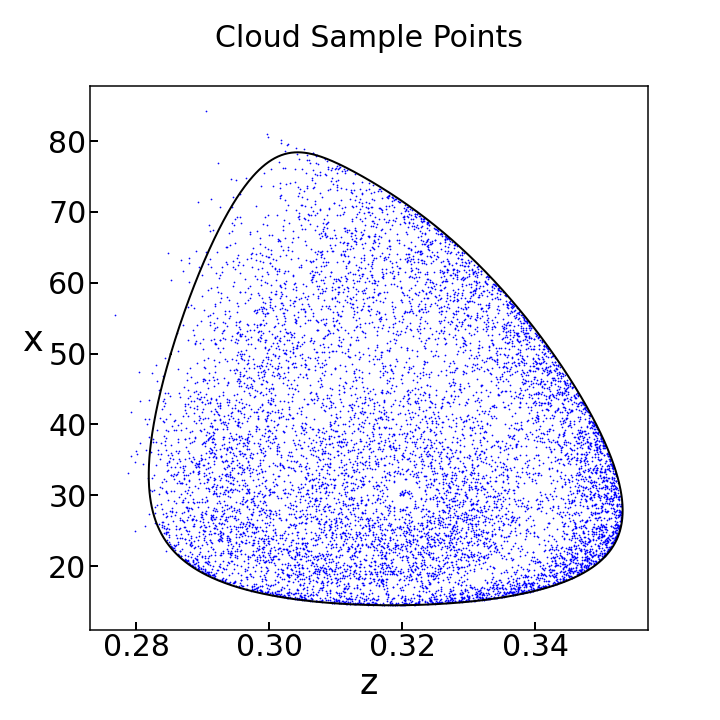}
        \caption{Sampling for Sections \ref{sec:RHSMetric} and \ref{sec:GBFunMetric}}
    \end{subfigure}
    \begin{subfigure}[b]{0.45\textwidth}
        \includegraphics[width=\textwidth]{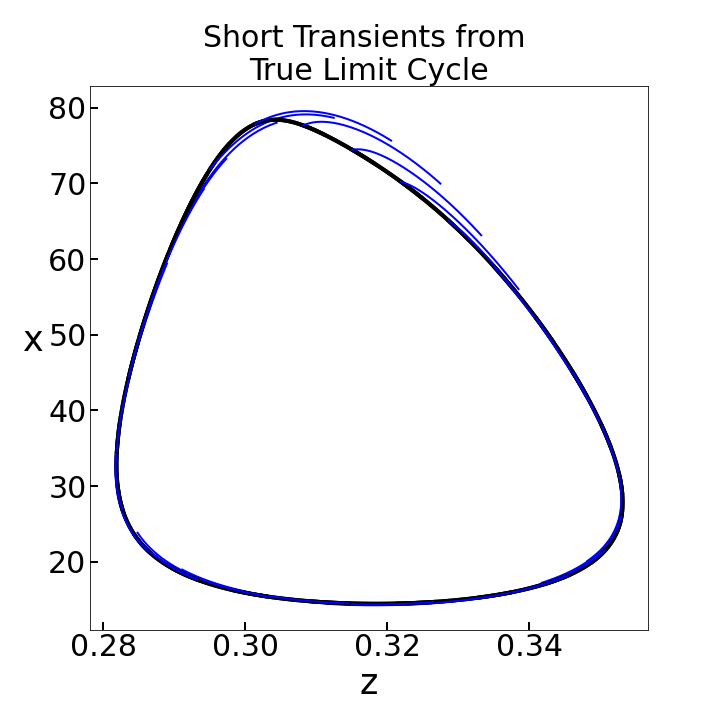}
        \caption{Short Term Transients for Section \ref{sec:HairMetric}}
    \end{subfigure}
    \caption{(a) Sampling points were obtained by randomizing initial conditions, integrating in time and discarding the early transients, and sub-sampling from 1000 different trajectories. (b) Short term transients obtained by taking as initial conditions 50 points evenly spaced around the true limit cycle and integrating for $t=20$}
    \label{fig:MetricsSampling}
\end{figure}

\subsection{Short-Term Prediction Error} \label{sec:HairMetric}

Let $\mathbf{x}_j^{LC}=\{x_i^{LC}\}_j$ be the steady-state solution if it is stable, and a point on the limit cycle if the steady-state solution is unstable. Let $N_v$ be the number of variables, $N_n$ the number of testing points, and $T$ the time period for short-term transients.\\

Starting from $\mathbf{x}_j^{LC}$, the true ODEs and the ANN learned ODEs are integrated for $t=T$.
\begin{equation}
\begin{aligned}
    \mathbf{x}_j(t=0) &= \mathbf{x}_j^{LC},\;\;1\leq j\leq N_n\\
    \mathbf{y}_j &= \int_0^T f(\mathbf{x},\mathbf{p})dt ,\;\;1\leq j\leq N_n\\
    \mathbf{\hat{y}}_j &= \begin{dcases}
    \int_0^T \mathcal{N}(\mathbf{x},\mathbf{p};\theta)dt ,\;\;1\leq j\leq N_n & (Black\;\;Box)\\
    \int_0^T g(\mathbf{x}, \mathbf{p}, \mathcal{N}(\mathbf{x},\mathbf{p};\theta))dt ,\;\;1\leq j\leq N_n & (Gray\;\;Box)
    \end{dcases}
\end{aligned}
\end{equation}

$\mathbf{y}_j$ and $\mathbf{\hat{y}}_j$ are the true and predicted matrices. These matrices are condensed into an error metric using various norms defined in section \ref{sec:norms}.

\subsection{RHS Error} \label{sec:RHSMetric}
\begin{equation}
\begin{aligned}
    \mathbf{y}_j &= f(\mathbf{x},\mathbf{p}),\;\;1\leq j\leq N_n\\
    \hat{\mathbf{y}}_j &= \begin{dcases}
    \mathcal{N}(\mathbf{x},\mathbf{p};\theta),\;\;1\leq j\leq N_n & (Black\;\;Box)\\
    g(\mathbf{x}, \mathbf{p}, \mathcal{N}(\mathbf{x},\mathbf{p};\theta)),\;\;1\leq j\leq N_n & (Gray\;\;Box)
    \end{dcases}
\end{aligned}
\end{equation}

$\mathbf{y}_j$ and $\mathbf{\hat{y}_j}$ are the true and predicted  matrices. These matrices are condensed into an error metric using various norms defined in section \ref{sec:norms}.

\subsection{Gray-Box Function Error} \label{sec:GBFunMetric}

For Gray-Box models, the neural network learns a subset of the full ODE function.
\begin{equation}
\begin{aligned}
    \dot{\mathbf{x}} &= f(\mathbf{x},\mathbf{p}) = g(\mathbf{x}, \mathbf{p}, \phi(\mathbf{x}, \mathbf{p})) && True\;\;ODE\\
    \hat{\dot{\mathbf{x}}} &= g(\mathbf{x}, \mathbf{p}, \hat{\phi}(\mathbf{x}, \mathbf{p})) && Learned\;\;ODE
\end{aligned}
\end{equation}

\begin{equation}
\begin{aligned}
    \mathbf{y}_j &= \phi(\mathbf{x},\mathbf{p}),\;\;1\leq j\leq N_n\\
    \hat{\mathbf{y}}_j &= \hat{\phi}(\mathbf{x},\mathbf{p}) = \mathcal{N}(\mathbf{x},\mathbf{p};\theta),\;\;1\leq j\leq N_n\\
\end{aligned}
\end{equation}

$\mathbf{y}_j$ and $\mathbf{\hat{y}_j}$ are the true and predicted matrices. These matrices are condensed into an error metric using various norms defined in section \ref{sec:norms}.

\subsection{Experimental Parameter Error}
Some Gray-Box models include additional parameters that have physical and experimental significance, and these are fit during ANN training. For these parameters, relative error is used.\\

Let $\boldsymbol{\kappa}=\{\kappa\}_k$ be the experimental parameters, with $\hat{\boldsymbol{\kappa}}=\{\hat{\kappa}\}_k$ the ANN predictions, and $N_P$ the number of learnable experimental parameters.
\begin{equation}
\begin{aligned}
    \mathcal{L}(\boldsymbol{\kappa},\hat{\boldsymbol{\kappa}}) &=
    \cfrac{1}{N_P}\sum_{k=1}^{N_P}\left|\cfrac{\kappa_k-\hat{\kappa}_k}{\kappa_k}\right|
\end{aligned}
\end{equation}

\subsection{Norms} \label{sec:norms}
The error matrices $\mathbf{y}_j$ and $\mathbf{\hat{y}}_j$ are first normalized with min-max scaling per variable. $\mathbf{\hat{y}}_j$ is scaled with the min/max of $\mathbf{y}_j$.
\begin{equation}
\begin{aligned}
    \tilde{\mathbf{y}}_j = \cfrac{\mathbf{y}_j-\min{(\mathbf{y})}}{\max{(\mathbf{y})}-\min{(\mathbf{y})}}\\
    \tilde{\hat{\mathbf{y}}}_j = \cfrac{\mathbf{y}_j-\min{(\hat{\mathbf{y})}}}{\max{(\mathbf{y})}-\min{(\mathbf{y})}}
\end{aligned}
\end{equation}

These normalized error matrices are condensed into a single metric using various norms:

\begin{equation}
\begin{aligned}
    \mathcal{L}_1(\tilde{\mathbf{y}}_j,\tilde{\hat{\mathbf{y}}}_j) &= \cfrac{1}{N_nN_V} \sum_{j=1}^{N_n}\sum_{i=1}^{N_V}|\tilde{y}_{i,j}-\tilde{\hat{y}}_{i,j}|\\
    \mathcal{L}_2(\tilde{\mathbf{y}}_j,\tilde{\hat{\mathbf{y}}}_j) &= \cfrac{1}{N_nN_V} \sum_{j=1}^{N_n}\sqrt{\sum_{i=}^{N_V}(\tilde{y}_{i,j}-\tilde{\hat{y}}_{i,j})^2}\\
    \mathcal{L}_\infty(\tilde{\mathbf{y}}_j,\tilde{\hat{\mathbf{y}}}_j) &=\cfrac{1}{N_n} \sum_{j=1}^{N_n}\max_i|\tilde{y}_{i,j}-\tilde{\hat{y}}_{i,j}|
\end{aligned}
\end{equation}

\end{document}